\documentclass[acmtocl,acmnow,hyperref]{vxh-acmtrans2m}

\usepackage{vxh-acm}

\begin{document}
\sloppy

\markboth{Xuan-Ha Vu \emph{et al.}}{Branch-and-Prune Search Strategies for
Numerical Constraint Solving}

\title{
    Branch-and-Prune Search Strategies\\
    for Numerical Constraint Solving
}

\author{
    Xuan-Ha VU$^1$,
    Marius-C\u{a}lin SILAGHI$^2$,
    Djamila SAM-HAROUD$^3$ and
    Boi FALTINGS$^3$\\
    \llap{$^1$\ }\href{http://4c.ucc.ie/}{Cork Constraint Computation Centre},
        \href{http://4c.ucc.ie/}{University College Cork} (UCC),\\
        14 Washington Street West, Cork, Ireland,\\
        Email: \href{mailto:ha.vu@4c.ucc.ie}{\texttt{ha.vu@4c.ucc.ie}};\\
    \llap{$^2$\ }\href{http://www.cs.fit.edu/}{Department of Computer Science},
        \href{http://www.fit.edu/}{Florida Institute of Technology} (FIT),\\
        150 West University Boulevard, Melbourne, FL-32901-6975, USA\\
        Email: \href{mailto:marius.silaghi@cs.fit.edu}{\texttt{marius.silaghi@cs.fit.edu}};\\
    \llap{$^3$\ }\href{http://liawww.epfl.ch/}{Artificial Intelligence Laboratory},
        \href{http://www.epfl.ch/}{Ecole Polytechniqe F\'ed\'eral de Lausanne} (EPFL),\\
        Batiment IN, Station 14, CH-1015 Lausanne, Switzerland,\\
        Email: \href{mailto:jamila.sam@epfl.ch}{\texttt{jamila.sam@epfl.ch}},
            \href{mailto:boi.faltings@epfl.ch}{\texttt{boi.faltings@epfl.ch}}.
}


\begin{abstract}

When solving \emph{numerical constraints} such as nonlinear equations and
inequalities, solvers often exploit \emph{pruning} techniques, which remove
redundant value combinations from the domains of variables, at \emph{pruning}
steps. To find the \emph{complete} solution set, most of these solvers
alternate the \emph{pruning} steps with \emph{branching} steps, which split
each problem into subproblems. This forms the so-called \emph{branch-and-prune}
framework, well known among the approaches for solving numerical constraints.
The basic branch-and-prune search strategy that uses \emph{domain bisections}
in place of the branching steps is called the \emph{bisection search}. In
general, the bisection search works well in case $(i)$ the solutions are
isolated, but it can be improved further in case $(ii)$ there are continuums of
solutions (this often occurs when inequalities are involved). In this paper, we
propose a new branch-and-prune search strategy along with several variants,
which not only allow yielding better branching decisions in the latter case,
but also work as well as the bisection search does in the former case. These
new search algorithms enable us to employ various pruning techniques in the
construction of inner and outer approximations of the solution set. Our
experiments show that these algorithms speed up the solving process often by
one order of magnitude or more when solving problems with continuums of
solutions, while keeping the same performance as the bisection search when the
solutions are isolated.

\end{abstract}





\renewcommand{\permission}{}

\begin{bottomstuff}
Preliminary parts of this paper have been published in
\cite{SilaghiMC-SH-F:2001,VuXH-SH-S:2002b,VuXH-SH-S:2002c}. This research was
mainly carried out at the Artificial Intelligence Laboratory, EPFL, and was
funded by the European Commission and the Swiss Federal Education and Science
Office through the COCONUT project (IST-2000-26063).
\end{bottomstuff}

\maketitle

\section{Introduction}
\label{sec:search-introduction}%

A \emph{constraint satisfaction problem} (CSP) consists of a finite set of
constraints specifying which value combinations from given \emph{domains} of
its variables are admitted. It is called a \emph{numerical constraint
satisfaction problem} (NCSP) if the domains are continuous. NCSPs such as
systems of equations and inequalities arise in many industrial applications. A
method for solving NCSPs is called \emph{complete} if every solution can be
found by it in the infinite time and can be approximated by it within an
arbitrarily small positive tolerance after a finite time, provided that the
underlying arithmetic is exact. Most available complete solution methods are
instances of the \emph{branch-and-prune} framework, which interleaves branching
steps with pruning steps. Roughly speaking, a \emph{branching} step divides a
problem into subproblems whose union is equivalent to the initial problem in
term of the solution set, and a \emph{pruning} step reduces a problem in some
measure. The reader can find a more detailed discussion on
\emph{branch-and-prune} methods in \cite[Section 3.2]{VuXH:Thesis:2005}.

The need for completeness arises in many applications such as safety
verification and computer-assisted proofs \cite[Section
3.1]{SchichlH:HThesis:2003}. In design applications such as estimation and
robust control \cite{AsarinE-B-O-D-O-P:2000,JaulinL-K-D-W:Book:2001},
automation and robotics
\cite{JaulinL-K-D-W:Book:2001,LeeE-M-M:2002:ASME,LeeE-M:2002,LeeE-M:2004,NeumaierA-M:2002},
civil engineering \cite{LottazC:Thesis:2000,VuXH:Thesis:2005}, and shape design
\cite{SnyderJM:Book:1992}, the solution set of an NCSP often expresses equally
relevant choices that need to be identified as precisely and as completely as
possible. In a design process, one often desires to find as many solutions as
possible because investigating many solutions at earlier stages potentially
increases the chance of success at later stages. This also allows identifying
good design choices. Thus, the complete solution set is often sought for,
provided that the response time is reasonable. Since a general NCSP is NP-hard,
the time for finding all solutions at a high precision is often prohibitively
long. In most cases, a low or medium precision is however sufficient for
applications. Hence, there is a tradeoff between timely but less precise
information and slow but more precise information.

The necessary background is presented in Section~\ref{sec:definition}. The rest
of the paper is organized as follows. We first prepare general settings about
solution representations and formal definitions of domain reduction and
splitting operators in Section~\ref{sec:search-rept-solutions} and
Section~\ref{sec:search-operators}, respectively. We then present, in
Section~\ref{sec:search-algorithm}, a generic branch-and-prune search
algorithm, called \ALG{BnPSearch}, that enables the incorporation of domain
reduction and splitting operators (Section~\ref{sec:bnp-search-algorithm}), and
then present, as instances of \ALG{BnPSearch}, the bisection search
(Section~\ref{sec:classic-search-algorithm}) and two new search algorithms,
called \ALG{UCA5} and \ALG{UCA6} (Section~\ref{sec:new-search-algorithm}).
Roughly speaking, the effectiveness of the new algorithms is mainly due to the
following policies, if some constraint, $C$, is predicted to contain continuums
of feasible points:
\begin{itemize}
    \item \textbf{Working on the negation of constraints to find feasible
    regions.} Apply \emph{domain reduction} techniques to the negation of $C$.
    Let $\IA{x}$ be the input domains and $\IA{x}'$ the resulting domains. Then
    $\IA{x}^* := \IA{x} \setminus \IA{x}'$ is feasible w.r.t. $C$. Thus, $C$
    can be immediately removed from subproblems defined on subregions of
    $\IA{x}^*$.

    \item \textbf{Reducing the number of variables in subproblems.} In a
    subproblem generated during the solving process, some variables may not
    occur in any constraint under consideration and thus no longer need to be
    considered.

    \item \textbf{Taking the influence of all constraints on each other into
    account at each iteration.} Potentially, this reduces the search
    space better than solving each constraint at a time does.
\end{itemize}

A simple heuristic to predict whether a constraint contains continuums of
solutions is to check if this constraint is an inequality. In general,
\ALG{UCA5} and \ALG{UCA6} allow better branching decisions than the basic
domain bisection. Although providing an accurate representation of solutions,
in some cases these algorithms are still slow and provide verbose
representations. The first reason is that the orthogonal splitting policy in
these algorithms generates a significant number of nearly aligned \emph{boxes}
near the boundaries of constraints. The second reason is that these algorithms
often have to spend too much effort on producing too small boxes with respect
to the precision $\varepsilon$, which is predetermined by users.

We later propose, in Section~\ref{sec:search-impr-algorithm}, an improved
instance, called \ALG{UCA6$^+$}, of \ALG{BnPSearch} to tackle the above two
limitations. The improvement in \ALG{UCA6$^+$} is twofold. First,
\ALG{UCA6$^+$} utilizes domain reduction techniques better when the precision
is recognized as being sufficient. Namely, it tells domain reduction techniques
to avoid unnecessarily spending too much effort on reducing the domains whose
sizes are smaller than $\varepsilon$. When the sizes of a certain number of
domains are smaller than $\varepsilon$, \ALG{UCA6$^+$} allows resorting to a
simple solver that is more efficient for small and very low dimensional
problems. The gain is then in both the computational time and the alignment of
boxes. Second, \ALG{UCA6$^+$} allows resorting to geometric representation
techniques to combine aligned boxes produced in the previous stage into larger
equivalent boxes. The representation of the solution set is therefore more
concise. This potentially accelerates querying and complicated operations on
the explicit representation of solutions.

In general, our new search algorithms improve the solving process when there
are continuums of solutions, and keep the same procedure and performance as the
bisection search when the solutions are isolated. In the former case, our
search algorithms allow producing \emph{inner} and \emph{outer approximations}
w.r.t. a predetermined precision $\varepsilon$. Moreover, a large percentage of
the outcome are often proved to be \emph{sound} solutions. Our experiments in
Section~\ref{sec:search-experiment} show that the new search algorithms
significantly improve the efficiency as well as the conciseness of solution
representations. The conclusion is finally given in
Section~\ref{sec:search-conclusion}.

\section{Background and Definition}
\label{sec:definition}%

\subsection{Numerical Constraint Satisfaction}

We recall in this section two central concepts of constraint programming.

\begin{definition}
\label{def:constraint}%
A \emph{constraint} on a finite sequence of variables, $\seq{x_1, \dots, x_k}$,
taking their values in respective domains, $\seq{D_1, \dots, D_k}$, is a subset
of the Cartesian product $D_1 \times \dots \times D_k$, where $k \in \N$ ($\N$
is the set of natural numbers).
\end{definition}

\begin{definition}[CSP]
\label{def:csp}%
A \emph{constraint satisfaction problem}, abbreviated to CSP, is a triple
$(\cl{V}, \cl{D}, \cl{C})$ in which $\cl{V}$ is a finite sequence of variables
$\seq{v_1, \dots, v_n}$, $\cl{D}$ is a finite sequence of the respective
domains of the variables $\seq{v_1, \dots, v_n}$, and $\cl{C}$ is a finite set
of constraints, each on a subsequence of $\cl{V}$. A \emph{solution} of this
problem is an assignment of values from $\cl{D}$ to $\cl{V}$ respectively such
that all constraints in $\cl{C}$ are satisfied. The set of all solutions is
called the \emph{solution set}.
\end{definition}

The reader can find many more concepts in \cite{AptKR:Book:2003}. In this
paper, we only focus on numerical CSPs, which are defined as follows.

\begin{definition}[NCSP]
\label{def:continuous-constraint}%
A domain is said to be \emph{continuous} if it is a real interval. A
\emph{numerical constraint} is a constraint on a sequence of variables whose
domains are continuous. If all constraints of a CSP are numerical, it is called
a \emph{numerical constraint satisfaction problem} (abbreviated to NCSP).
\end{definition}

An NCSP can be viewed as a constrained optimization problem with a constant
objective function. Thus, it can be theoretically solved by using
\emph{mathematical programming} (MP) methods for solving constrained
optimization problems. However, most of the efficient MP methods are heavily
based on the influence between objective functions and constraints, thus not
efficient for directly solving NCSPs.

Since thirty years ago, \emph{constraint satisfaction} techniques have been
being devised to solve CSPs with discrete domains. These techniques perform
\emph{reasoning} procedures on constraints and explore the search space by
intelligently enumerating solutions. In order to solve NCSPs by means of
constraint satisfaction, continuous domains have often been converted into
discrete domains by using progressive \emph{discretization} techniques
\cite{SamHaroudD:Thesis:1995,LottazC:Thesis:2000}. Still, these methods are
often inefficient. Later on, many mathematical \emph{computing} techniques for
continuous domains have been integrated into the framework of constraint
satisfaction in order to solve NCSPs more efficiently. Nowadays, these
techniques are often referred to as \emph{constraint programming} techniques,
which imply the combination of \emph{computing} and \emph{reasoning} aspects. A
much more extensive discussion on numerical constraint solving can be found in
\cite{VuXH:Thesis:2005}.

\subsection{Interval Arithmetic}
\label{sec:interval}

Let $\setinf{\R} \equiv \R \cup \set{-\infty, +\infty}$. The \emph{lower bound}
of a real interval $\IA{x}$ is defined as $\inf(\IA{x}) \in \setinf{\R}$, and
the \emph{upper bound} of $\IA{x}$ is defined as $\sup(\IA{x}) \in
\setinf{\R}$. Let denote $\Lb{x} = \inf(\IA{x})$ and $\Ub{x} = \sup(\IA{x})$.
There are four possible intervals $\IA{x}$ with these bounds:
\pagebreak[1]%
{\allowdisplaybreaks%
\begin{eqnarray*}
\renewcommand{\arraystretch}{1.1}
\setlength{\arraycolsep}{4pt}
\hspace{-0.5em}%
\begin{array}{llllllll}
    \bullet & \tn{The \emph{closed interval} defined as} && \IA{x} & \equiv &
    \cintv{\Lb{x}}{\Ub{x}} & \equiv & \set{x \in \R \mid \Lb{x} \le x \le
    \Ub{x}};
    \\
    \bullet & \tn{The \emph{open interval} defined as} && \IA{x} & \equiv &
    \ointv{\Lb{x}}{\Ub{x}} & \equiv & \set{x \in \R \mid \Lb{x} < x < \Ub{x}};
    \\
    \bullet & \tn{The \emph{left-open interval} defined as} && \IA{x} & \equiv
    & \lintv{\Lb{x}}{\Ub{x}} & \equiv & \set{x \in \R \mid \Lb{x} < x \le
    \Ub{x}};
    \\
    \bullet & \tn{The \emph{right-open interval} defined as} && \IA{x} & \equiv
    & \rintv{\Lb{x}}{\Ub{x}} & \equiv & \set{x \in \R \mid \Lb{x} \le x <
    \Ub{x}}.
\end{array}
\end{eqnarray*}
}%
Let $\I$ be the set of all closed intervals and $\II$ the set of all intervals.
The \emph{interval hull} of a subset $S$ of $\R$ is the smallest interval
(w.r.t. the set inclusion), denoted as $\Hull{S}$, that contains $S$. For
example, $\Hull{(\lintv{1}{3} \cup \set{2, 4})} = \lintv{1}{4}$. Given a
nonempty interval $\IA{x}$, we define that the \emph{midpoint} of $\IA{x}$ is
$\Mid(\IA{x}) \equiv (\inf(\IA{x}) + \sup(\IA{x}))/2$ and the \emph{width} of
$\IA{x}$ is $\Wid(\IA{x}) \equiv {\sup(\IA{x}) - \inf(\IA{x})}$. We also agree
that $\Wid(\emptyset) = 0$ and $\Mid(\emptyset) = 0$. A \emph{box} is the
Cartesian product of a finite number of intervals. The concepts of the midpoint
and width are defined on \emph{boxes} in a componentwise manner.

Fundamentally, if $\IA{x}$ and $\IA{y}$ are two (real) intervals, then the four
elementary operations for \emph{idealized interval arithmetic} obey the rule
\begin{equation}
\label{eqn:ia}
    \IA{x} \diamond \IA{y} \equiv \set{x \diamond y \mid x \in \IA{x}, y \in
    \IA{y}}, \quad \forall \diamond \in \set{+, -, *, \div}.
\end{equation}
Thus, the results of the four elementary operations in interval arithmetic are
exactly the ranges of their real-valued counterparts. Although the rule
\eqref{eqn:ia} characterizes these operations mathematically, the usefulness of
interval arithmetic is due to the \emph{operational definitions} based on
interval bounds. For example, let $\IA{x} = \intv{x}$ and $\IA{y} = \intv{y}$
be two closed intervals, interval arithmetic shows that:
\pagebreak[0]%
\begin{eqnarray}
\label{eqn:ia-fund-operation}%
    \IA{x} + \IA{y} & \equiv & \cintv{\Lb{x}+\Lb{y}}{\Ub{x}+\Ub{y}};
    \\
    \IA{x} - \IA{y} & \equiv & \cintv{\Lb{x}-\Ub{y}}{\Ub{x}-\Lb{y}};
    \\
    \IA{x}\, * \, \IA{y} & \equiv & \cintv{
    \min \set{\Lb{x}\Lb{y}, \Lb{x}\Ub{y}, \Ub{x}\Lb{y}, \Ub{x}\Ub{y}}}
    {\max \set{\Lb{x}\Lb{y}, \Lb{x}\Ub{y}, \Ub{x}\Lb{y}, \Ub{x}\Ub{y}}};
    \\
    \IA{x} \div \IA{y} & \equiv & \IA{x} * (1/\IA{y}) \tn{ if } 0 \notin
    \IA{y}, \tn{ where } 1/\IA{y} \equiv  \cintv{1/\Ub{y}}{1/\Lb{y}}.
\end{eqnarray}
Simple \emph{arithmetic expressions} are composed of these four elementary
operations.

An \emph{interval form}, $\IA{f} : \II^m \to \II^n$, of a real function $f : D
\subseteq \R^m \to \R^n$ is constructed conforming to the \emph{inclusion
property}: the value of the interval form encloses the exact range of the real
function, that is, $\forall \IA{x} \in \II^m : f(\IA{x}) \subseteq
\IA{f}(\IA{x})$ or, equivalently, $\forall \IA{x} \in \II^m, x \in D : x \in
\IA{x} \OIF f(x) \in \IA{f}(\IA{x})$.

The finite nature of computers precludes an exact representation of the real
numbers. The real set $\R$ is therefore approximated by a finite set $\F$ of
\emph{floating-point numbers} \cite{GoldbergD:1991}, including $-\infty$ and
$+\infty$. The set of real intervals is then replaced with the set $\FI$ of
closed \emph{floating-point intervals} with bounds in $\F$. The interval
concepts are similarly defined on $\FI$ while conforming to the inclusion
property. The power of interval arithmetic lies in its implementation on
computers. In particular, \emph{outwardly rounded} interval arithmetic allows
computing \emph{rigorous enclosures} of the ranges of functions. An interval is
said to be \emph{canonical} iff it does not contain two different intervals
whose union is not an interval. A box is said to be \emph{canonical} iff all
its intervals are canonical.

The reader can find extended introductions to interval analysis in
\cite{MooreRE:Book:1966,MooreRE:Book:1979,AlefeldG-H:Book:1983}, interval
methods for systems of equations in \cite{NeumaierA:Book:1990}, interval
methods for optimization in \cite{HansenER-W:Book:2004}, and some recent
applications of interval arithmetic in \cite{JaulinL-K-D-W:Book:2001}.

Most interval arithmetic libraries have been implemented for closed intervals
only. One however can use these libraries to perform some computations on
open/closed intervals. Indeed, every interval $\IA{x}$ with bound values
$\Lb{x}$ and $\Ub{x}$ is contained in the corresponding closed interval
$\intv{x} \in \I$. If the computations are domain reduction or complementary
boxing, we can perform these computations on the corresponding closed intervals
of the domains to get new domains, and then take the set intersection of these
new domains with the initial general intervals. For example, after performing a
domain reduction technique on the closed interval $\intv{x}$, we get a closed
interval $\intv{y} \subseteq \intv{x}$. The result to be obtained is the set
intersection $\IA{x} \cap \intv{y}$, thus may be open or closed.

\subsection{A Short Overview of Branch-and-Prune Solution Methods}
\label{sec:overview-bnp}%

To be able to find the \emph{complete} solution set of an NCSP, most solvers
follow the \emph{branch-and-prune} framework, a well known approach for solving
numerical constraints. In this framework, a solver alternates pruning steps
with branching steps until reaching the required precision, where a
\emph{pruning} step attempts to reduce each considered problem in some measure
and a \emph{branching} step splits each considered problem into subproblems.
The basic branch-and-prune search that uses \emph{domain bisections} in place
of the branching steps is called the \emph{bisection search}.

A pruning technique that attempts to reduce the domains of problems without
discarding any solution is called a \emph{domain reduction} technique. In
contrast to a domain reduction technique, a \emph{test} (e.g., an existence
test, a uniqueness test, an exclusion test and an inclusion test) does not
change the domains of an input problem but maps this problem to a predetermined
status. In particular, \emph{existence}, \emph{uniqueness} and \emph{exclusion}
tests are to check if a problem has at least one solution, a unique solution
and no solution, respectively. The outcome of these tests is thus a Boolean
value. Quite differently, an \emph{inclusion test} is to check if all points
under consideration are solutions, non-solutions, or else; which is defined as
follows.

\begin{definition}
\label{def:inclusion-test}%
Let $X$ be a sequence of $n$ real variables. An \emph{inclusion test} is a
function $\FC$ that takes as input a (domain) box $\IA{x} \in \II^n$ and a
finite set $\cl{C}$ of constraints on a subsequence $Y$ of $X$ (assuming $Y$ is
well defined on $\IA{x}$) and that returns either \CONST{feasible},
\CONST{infeasible}, or \CONST{unknown} such that:
\begin{enumerate}
    \item If $\FC(\IA{x}, \cl{C}) = \CONST{feasible}$, then every point in
    $\IA{x}$ satisfies all constraints in $\cl{C}$.

    \item If $\FC(\IA{x}, \cl{C}) = \CONST{infeasible}$, then no point in
    $\IA{x}$ satisfies all constraints in $\cl{C}$.
\end{enumerate}
The inclusion test $\FC$ is said to be \emph{trivial} if it always returns
$\CONST{unknown}$. It is said to be \emph{$\varepsilon$-strong} if, for every
$\IA{x}$, the truth of $\Wid(\IA{x}) \le \varepsilon \wedge \FC(\IA{x}, \cl{C})
= \CONST{unknown}$ implies the existence of a feasible point of $\cl{C}$ in
$\IA{x}$.
\end{definition}

\begin{figure}[!htb]
\centering
    \includegraphics[width=9cm]{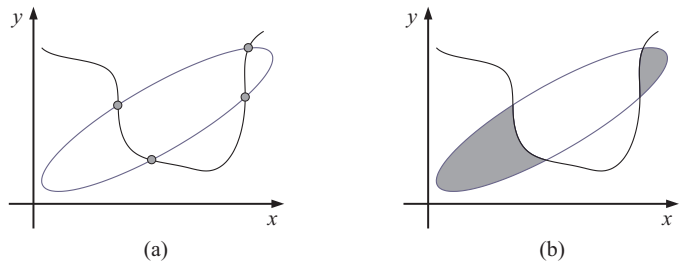}
\caption[An example of NCSPs with: (a) isolated solutions; (b) continuums of
solutions]{(a) An NCSP with four isolated solutions (grey dots); (b) An NCSP
with two continuums of solutions (grey regions).}
\label{fig:continuum}%
\end{figure}

An extended overview of fundamental and recent complete methods for solving
NCSPs has been presented in \cite[Chapter~3]{VuXH:Thesis:2005}. In summary,
those methods can be viewed as instances of the branch-and-prune framework.
Many of them integrate existence, uniqueness or exclusion tests at pruning
steps of the bisection search. At a branching step, they bisect a domain, which
is often chosen as the largest with respect to some measure (e.g., domain
size), provided that this domain is amenable to be split (e.g, its size is
greater than a predetermined precision $\varepsilon$). Because most solution
methods have been designed to solve a square system of equations, they only aim
at generating a collection of tiny boxes, each encloses a solution. This
approach is referred to as the \emph{point-wise} approach. It may be reasonable
for solving NCSPs with \emph{isolated solutions} (see
Figure~\ref{fig:continuum}a), but are often inefficient, when applied in a
straightforward manner, for solving NCSPs with \emph{continuums of solutions}
(see Figure~\ref{fig:continuum}b). In the latter case, neither the
computational time nor the compactness of the solution representation are
satisfactory.

It is possible to enhance the solving process in the point-wise approach by
replacing the existence, uniqueness and exclusion tests with domain reduction
techniques in a straightforward manner. In the rest, the resulting search
strategy will be called \emph{dichotomous maintaining bounds by consistency}
(\ALG{DMBC}). When solving NCSPs with non-isolated solutions, \ALG{DMBC} search
techniques may be able to cover a spectrum of non-isolated solutions with a
number of subsets of $\R^n$. However, it is often not possible to prove that a
subset of the outcome are all solutions.

In contrast to the point-wise approach, another one, called the
\emph{set-covering} approach, has been developed in order to represent
continuums of solutions more reasonably. It aims at covering, as accurately as
possible, continuums of solutions with inner and outer approximations, each
consists of a number of subsets of $\R^n$. All points represented by the inner
approximation are proved to be solutions. Usually, the representation of inner
and outer approximations is made simple such that the costs of usual operations
on this representation are as cheap as possible. The subsets in these
approximations are often chosen to be simple ones (e.g., boxes). Owning to
their nice properties, boxes have been used in many set-covering techniques,
thus forming the so-called \emph{box-covering} techniques.

\emph{\textbf{For simplicity, in this paper we restrict our attention to
solution methods that use boxes as elements of the outcome.}} One has often
employed inclusion tests at pruning steps of typical box-covering
branch-and-prune methods to prove that all points in some subsets are
solutions. In this paper, a \ALG{DMBC}-like search technique that combines both
domain reduction techniques and inclusion tests at its pruning steps is called
a \ALG{DMBC$^+$} search technique (see
Section~\ref{sec:classic-search-algorithm}). Therefore, \ALG{DMBC$^+$} search
techniques may, depending on the strong of employed inclusion tests, be able to
provide inner approximations of the solution set. We hence say that
\ALG{DMBC$^+$} search techniques belong to the box-covering approach and that
\ALG{DMBC} search techniques belong to the point-wise approach. Note that both
\ALG{DMBC} and \ALG{DMBC$^+$} search strategies do not remove constraints from
consideration during the solving process. That is, they always consider the
whole set of initial constraints when resorting to domain reduction techniques
and tests.

When solving an NCSP with continuums of solutions at a high precision,
\ALG{DMBC} techniques often provide a huge collection of tiny boxes as an outer
approximation of the solution set while \ALG{DMBC$^+$} techniques are usually
able to provide both inner and outer approximations of the solution set, each
approximation is a much more concise collection of boxes. However, the number
of boxes used by \ALG{DMBC$^+$} techniques to approximate the boundary of
continuums of solutions is still very high in many cases. Therefore, either
their applicability is restricted or the tractability limit is rapidly reached.
The new search techniques proposed in this paper will tackle this limitation
(see Section~\ref{sec:search-algorithm} and
Section~\ref{sec:search-impr-algorithm}).

Because most computers were equiped with floating-point number systems, an
important issue is how to deal with rounding errors occurring in computations
on floating-point numbers. Owning to the inclusion property of outwardly
rounded interval arithmetic, one has often been using it to tackle the issue of
rounding errors. Recent interval arithmetic based methods are able to solve a
number of NCSPs efficiently while still enjoying the completeness. Those
methods can be viewed as instances of the branch-and-prune framework, although
they mainly focus on improving the pruning steps.

Interval constraint solvers such as \PRO{CLP}(BNR)
\cite{BenhamouF-O:1992,BenhamouF-O:1997}, \PRO{Numerica}
\cite{VanHentenryckP:1998} and \PRO{ILOG Solver} \cite{ILOGSolver-6.0:2003}
have shown their ability to efficiently find all solutions of certain instances
of NCSPs within an arbitrary positive tolerance. They are instances of the
branch-and-prune framework, and most of them use a simple branching policy like
the domain bisection or dichotomization as their default branching policy while
leaving more advanced branching policies to users. In other words, they
essentially follow the point-wise approach and only aim at solving NCSPs with
isolated solutions.

Recently, a number of box-covering methods have been developed in
\cite{JaulinL:Thesis:1994,JaulinL-K-D-W:Book:2001}, \cite{SamHaroudD-F:1996},
\cite{GarloffJ-G:1999}, \cite{CollavizzaH-D-R:1999} and
\cite{BenhamouF-G:2000,BenhamouF-G-L-C:2004} in order to represent continuums
of solutions. Although those methods are more suitable for dealing with
continuums of solutions than the point-wise methods are, their branching
policies still have at least one of the following limitations:
\begin{itemize}
    \item \textbf{They are not complete methods in general.} For example,
    \citeN{CollavizzaH-D-R:1999} have proposed a technique to extend a known
    feasible box of an inequality of the form $f(x) \le 0$ by performing box
    consistency (a kind of domain reduction) on its \emph{associated equation},
    $f(x) = 0$. Unfortunately, their results (Proposition~1 and 2 in that
    paper) do not hold for general constraints.

    \item \textbf{They are only designed for special constraints.} For example,
    the technique proposed in \cite{GarloffJ-G:1999} uses Bernstein polynomials
    to construct algebraic inclusion tests for use in a
    \ALG{DMBC}/\ALG{DMBC$^+$} search, and is restricted to polynomial
    constraints. The technique proposed in
    \cite{SamHaroudD-F:1996,LottazC:Thesis:2000} is restricted to the class of
    NCSPs with convexity properties. The technique proposed in
    \cite{BenhamouF-G:2000,BenhamouF-G-L-C:2004} is originally designed for
    solving universally quantified constraints.

    \item \textbf{They do not fully exploit the power of domain reduction
    techniques.} Namely, they only interleave \emph{inclusion tests} with
    uniformly splitting policies on all variables: each box produced by the
    splitting is tested for inclusion. The outcome can be structured into the
    form of a \emph{$2^k$-tree}. In \cite{JaulinL:Thesis:1994}, this process is
    performed in the space of all variables. However, in
    \cite{SamHaroudD-F:1996,LottazC:Thesis:2000}, only binary and ternary
    constraints obtained by \emph{ternarizing}\footnote{Ternarizing an NCSP is
    to recursively replace each binary arithmetic subexpression with an
    auxiliary variable until the arity of all the resulting expressions is at
    most three.} the initial NCSP are considered for the construction of
    $2^k$-trees.

    \item \textbf{They do not sufficiently take the influence of constraints on
    each other into account during the solving process.} For example, the
    solution method in \cite{SamHaroudD-F:1996,LottazC:Thesis:2000} construct
    \emph{quadtrees} and \emph{octrees} for individual binary and ternary
    constraints, respectively, and finally perform a constraint propagation on
    those trees. On the other hand, at each iteration the solution method in
    \cite{BenhamouF-G:2000,BenhamouF-G-L-C:2004} considers a constraint and
    solves it within every (domain) box produced as the outcome of
    the previous iteration.
\end{itemize}

Most methods represent the solution set of an NCSP \emph{explicitly} in the
space of initial variables, thus suffering from the high space complexity when
there are continuums of solutions. The only one exception we know of is the
work of \citeN{SamHaroudD-F:1996} and \citeN{LottazC:Thesis:2000}, in which the
authors have proposed to replace the explicit representation (in the space of
initial variables) with a \emph{semi-explicit} one, which is maintained by a
number of \emph{quadtrees} and \emph{octrees}. Although that semi-explicit
representation reduces the space complexity, it increases the querying time for
a solution. Notice that the uniformity of splitting must be maintained in those
methods in order to do propagation among $2^k$-trees. Therefore, the power of
domain reduction cannot be fully exploited in this method.

One of the most recent improvements to search is the work in
\cite{BenhamouF-G:2000,BenhamouF-G-L-C:2004}, which is summarized as follows.
In order to find feasible regions of universally quantified constraints of the
form $\forall t \in D_t: f(x, t) \le 0$,
\citeN{BenhamouF-G:2000,BenhamouF-G-L-C:2004} have proposed to perform a kind
of domain reduction on their negation, $f(x, t) > 0$. This operation is called
a \emph{negation test}. It encloses all possibly infeasible regions and the
remaining is feasible. Now consider a subproblem living in a domain box
$\IA{x}$. A procedure, called \ALG{ICAb3$_c$} in \cite{BenhamouF-G:2000}, takes
as input a constraint, $C$. The procedure \ALG{ICAb3$_c$} performs a negation
test on $C$ to reduce $\IA{x}$ to a new domain $\IA{x}'$. If $\IA{x}'$ is an
empty set, then every point in $\IA{x}$ satisfies $C$; otherwise, split $\IA{x}
\setminus \IA{x}'$ into boxes and then dichotomize $\IA{x}'$. Now, $C$ can be
removed from all subproblems that do not have domains in $\IA{x}'$. The
procedure \ALG{ICAb3$_c$} recursively performs the above operations on all
resulting subproblems that still have $C$ as a \emph{running
constraint}\footnote{A running constraint is a constraint that is currently
still under consideration.} and that have domains larger than a predetermined
precision $\varepsilon$. A search method, called \ALG{ICAb5} in
\cite{BenhamouF-G:2000}, repeats the procedure \ALG{ICAb3$_c$} for each
constraint, one by one, until all constraints have been processed. See also
Section~\ref{sec:search-cb-operators} for more discussion on the negation-based
approach.

\subsection{Representation of Non-isolated Solutions}
\label{sec:search-rept-solutions}

\subsubsection{Inner and Outer Approximations}

In case the solution set of an NCSP is empty or consists of isolated points,
its representation is usually simple. The representation of the solution set is
not simple in other cases, especially when the solution set contains continuums
of solutions. In general, the solution set of an NCSP is a relation on $\R^n$,
where $n$ is the number of variables in the NCSP. A relation can be
theoretically approximated by a superset and/or a subset.

\begin{definition}[Inner Approximation]
\label{def:inner-approx}%
Given a relation, $S \subseteq \R^n$, a set $S^{-} \subseteq \R^n$ is called an
\emph{inner approximation} of $S$ if it is contained in $S$; that is, $S^{-}
\subseteq S$.
\end{definition}

\begin{definition}[Outer Approximation]
\label{def:outer-approx}%
Given a relation, $S \subseteq \R^n$, a set $S^{+} \subseteq \R^n$ is called an
\emph{outer approximation} of $S$ if it contains $S$; that is, $S^{+} \supseteq
S$.
\end{definition}

When a relation on $\R^n$, such as the solution set of an NCSP, is approximated
by an inner approximation and/or an outer approximation. The latter is a
\emph{sound approximation} (i.e., it only contains solutions), but may lose
some solutions. Conversely, the former is a \emph{complete approximation}
(i.e., it contains all solutions), but may contain some points that are not
solutions. Given an exact representation $\cl{R}$, such as a collection of
boxes or a tree of boxes, of a relation $S \subseteq \R^n$, we denote by
$\pts(\cl{R})$ the set of points in $S$.

One often uses the volume difference, $\Vol(S^{+}) \setminus \Vol(S^{-})$, to
measure the degree of mismatch between inner and outer approximations. The
exact approximation errors are then bounded by this measure.

\subsubsection{Union Approximations}

Since the time for querying a point in a box is constant, one often
approximates a relation $S \subseteq \R^n$ by a collection of \emph{pairwise
disjoint boxes}, where two boxes are said to be (strictly)\footnote{The main
results in this paper still hold if we relax the disjointness such that
disjoint boxes may have common points on their facets but not in their
interiors.} \emph{disjoint} if they have no common points. Such a collection is
called a \emph{collection of disjoint boxes} for short. The representation of a
collection of disjoint boxes which is constructed by enumerating these boxes
and storing their coordinates is called the \emph{disjoint box representation}
\cite{AguileraA:Thesis:1998}. Among the approximations by collections of boxes,
the following three attract the most attention in practice because of their
simplicity.

\begin{definition}
\label{def:inner-union-approx}%
Given a relation $S \subseteq \R^n$. An \emph{inner union approximation} of
$S$, denoted by $\UNION[S]{I}$, is a collection of disjoint boxes in $\II^n$
such that $S \supseteq \pts(\UNION[S]{I})$.
\end{definition}

\begin{definition}
\label{def:outer-union-approx}%
Given a relation $S \subseteq \R^n$. An \emph{outer union approximation} of
$S$, denoted by $\UNION[S]{O}$, is a collection of disjoint boxes in $\II^n$
such that $S \subseteq \pts(\UNION[S]{O})$.
\end{definition}

\begin{definition}
\label{def:boundary-union-approx}%
Given a relation $S \subseteq \R^n$. A \emph{boundary union approximation},
denoted by $\UNION[S]{B}$, of $S$ (with respect to an inner union approximation
$\UNION[S]{I}$ and an outer union approximation $\UNION[S]{O}$) is a collection
of disjoint boxes in $\II^n$ such that $\pts(\UNION[S]{B}) = \pts(\UNION[S]{O})
\setminus \pts(\UNION[S]{I})$.
\end{definition}

\begin{remark}
Notice that $\UNION{X}$ is not a function, where $\cl{X} \in \set{\cl{I},
\cl{O}, \cl{B}}$. In this paper, we will always refer to $\UNION[S]{B}$ with
respect to some $\UNION[S]{I}$ and some $\UNION[S]{O}$, even when not mentioned
explicitly.
\end{remark}

\begin{figure}[!tb]
    \centerline{\includegraphics[width=3.8cm]{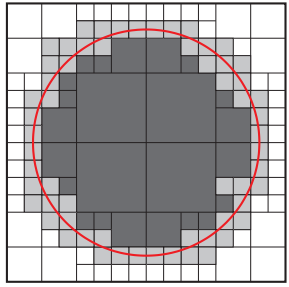}}
\caption[An example of inner and outer union approximations]{An example of
inner/outer/boundary union approximations of a circle with interior: the
collection of the dark grey boxes is an inner union approximation (\UNION{I});
the collection of the light grey boxes is a boundary union approximation
(\UNION{B}); the collection of the light and dark grey boxes is an outer union
approximation (\UNION{O}).}
\label{fig:inner-outer-union}%
\end{figure}

The concepts of union approximations are depicted in
Figure~\ref{fig:inner-outer-union}. Note that we always have the identity
$\pts(\UNION[S]{I}) \cap \pts(\UNION[S]{B}) = \emptyset$. In practice, one
often computes $\UNION[S]{I}$ and $\UNION[S]{B}$ first, and then obtains
$\UNION[S]{O}$ simply by $\UNION[S]{O} := \UNION[S]{I} \cup \UNION[S]{B}$.

The worst-case query time of a \emph{bounding-box tree} in $\R^d$ is
$\Theta(N^{1-1/d} + k)$, where $N$ is the number of boxes and $k$ is the number
of boxes intersecting the \emph{query range} \cite{AgarwalPK-B-G-H-H:2001}. It
is therefore useful to construct inner and/or outer union approximations of an
unknown relation (e.g., the solution set of an NCSP) in the form of a
\emph{bounding-box tree}. That is, the box represented by any node of the tree
contains the box represented by its child node, and all boxes represented by
the children of any node are disjoint. Fortunately, this property is enjoyed by
branch-and-prune solution methods assuming that the domains are intervals.

Several authors have recently addressed the construction of inner and/or outer
union approximations of the solution set of an NCSP. In
\cite{JaulinL:Thesis:1994}, union approximations are hierarchically constructed
in the form of a \emph{$2^k$-tree} in the space of initial variables. This
technique has shown its practical usefulness in robotics, automation and robust
control. The method in \cite{SamHaroudD-F:1996,LottazC:Thesis:2000} also aims
at the construction of $2^k$-trees in a similar way. However, only binary and
ternary constraints obtained by ternarizing the initial NCSP are considered for
the construction. That is, only \emph{quadtrees} and \emph{octrees} are
constructed. The solution set is finally approximated by a number of
\emph{quadtrees} and \emph{octrees} rather than a single $2^k$-tree. The space
complexity of approximations is thus reduced. The approach is however
restricted to the class of NCSPs with convexity properties.

Most recently, the method proposed in
\cite{BenhamouF-G:2000,BenhamouF-G-L-C:2004} corrects and extends the idea in
\cite{CollavizzaH-D-R:1999} to construct inner union approximations of
universally quantified constraints. Namely, when solving a universally
quantified constraint of the form $\forall t \in D_t : f(x, t) \le 0$, the
application of a domain reduction technique on the constraint $f(x, t)
> 0$ allows finding feasible regions on which an efficient search is based (see
Section~\ref{sec:search-cb-operators}).

\subsubsection{The Precision and Accuracy of Union Approximations}
\label{sec:qualification-fc}

The cost for achieving a given accuracy of approximations is often very high.
Alternatively, most constraint solvers stop splitting a box, which represents
the domains of a subproblem, as soon as the size of this box is not greater
than a given positive precision $\varepsilon$ (and this box is called an
\emph{$\varepsilon$-bounded box}). Some other solvers may attempt to apply a
pruning technique or a test to $\varepsilon$-bounded boxes before classifying
them as \emph{undiscernible}; thus, the name \emph{undiscernible box} has come
out.

In general, different constraint solvers use different criteria for leaving
$\varepsilon$-bounded boxes unprocessed. If a technique that is applied to
$\varepsilon$-bounded boxes before claiming them as undiscernible is used by a
solver, then it can be used for the other solvers as well. Therefore, the
comparison of search techniques should be based on the same criteria of
classifying $\varepsilon$-bounded boxes as \emph{undiscernible}. We propose to
use monotonic inclusion tests defined in the following definition for this
purpose. Let $\proj{\IA{x}}{Y}$ denote the \emph{projection} of a set $\IA{x}$
on the subsequence $Y$ of variables.

\begin{definition}[Monotonicity]
\label{def:fc-checker}%
Let use the same notations as in Definition~\ref{def:inclusion-test}. The
inclusion test $\FC$ is said to be \emph{monotonic} if, for every box $\IA{x}'$
and every finite set $\cl{C}'$ of constraints on $Y$ such that
$\proj{\IA{x}}{Y} \subseteq \proj{\IA{x}'}{Y}\subseteq \bigcap_{C \in \cl{C}'}
C$, we have
\begin{equation}
    \FC(\IA{x}, \cl{C}) = \CONST{unknown}\ \ \OIF\ \ \FC(\IA{x}', \cl{C}) =
    \FC(\IA{x}', \cl{C} \cup \cl{C}') = \CONST{unknown}.
\end{equation}
\end{definition}

Once the monotonicity holds for the value $\CONST{unknown}$ of an inclusion
test, it also holds for other values as shown below.

\begin{theorem}
\label{thm:fc-mononicity}%
Let use the same notations as in Definition~\ref{def:fc-checker}. If $\FC$ is a
monotonic inclusion test, then
\begin{itemize}
    \item If $\FC(\IA{x}, \cl{C}) = \CONST{feasible}$ holds, then $\FC(\IA{x}',
    \cl{C}) = \FC(\IA{x}', \cl{C} \cup \cl{C}') = \CONST{feasible}$ holds for
    every finite set $\cl{C}'$ of constraints on $Y$ and every nonempty box
    $\IA{x}'$ such that $\proj{\IA{x}'}{Y} \subseteq \proj{\IA{x}}{Y} \subseteq
    \bigcap_{C \in \cl{C}'} C$ holds;

    \item If $\FC(\IA{x}, \cl{C}) = \CONST{infeasible}$ holds, then $\FC(\IA{x}',
    \cl{C}) = \FC(\IA{x}', \cl{C} \cup \cl{C}') = \CONST{infeasible}$ holds for
    every finite set $\cl{C'}$ of constraints on $Y$ and every box $\IA{x}'$
    such that $\proj{\IA{x}'}{Y} \subseteq \proj{\IA{x}}{Y} \subseteq
    \bigcap_{C \in \cl{C}'} C$ holds.
\end{itemize}
\end{theorem}

\begin{proof}
This is easily proved by contradiction based on
Definition~\ref{def:fc-checker}.
\end{proof}

Based on the concept of a monotonic inclusion test, we define the precision of
union approximations (or a solution algorithm computing them) as follows.

\begin{definition}
\label{def:interval-precision}%
Given an NCSP $\cl{P} = (\cl{V}, \cl{D}, \cl{C})$, a precision (vector)
$\varepsilon > 0$, and a monotonic inclusion test $\FC$. A solution algorithm
that computes inner and boundary union approximations is (and thus those
approximations are) said to be of \emph{the precision $\varepsilon$ w.r.t. the
monotonic inclusion test $\FC$} if the boundary union approximation equals
(w.r.t. the set union) to a collection $\cl{U}$ of disjoint
$\varepsilon$-bounded or canonical boxes in $\II^n$ such that
\begin{equation}
    \forall \IA{x} \in \cl{U} : \FC(\IA{x}, \cl{C}) = \CONST{unknown}.
\end{equation}
If $\FC$ is trivial, we say for short that the solution algorithm and the
computed approximations are of \emph{the precision $\varepsilon$}. If $\FC$ is
$\varepsilon$-strong, we say that the solution algorithm and the computed
approximations are \emph{$\varepsilon$-accurate}.
\end{definition}

It is easy to see that a solution algorithm is complete if it is
$\varepsilon$-accurate (i.e., $\FC$ is $\varepsilon$-strong) for all
sufficiently small $\varepsilon
> 0$.

\section{Reduction and Splitting Operators for Exhaustive Search}
\label{sec:search-operators}%

Our improvements to the classic search strategies (i.e., \ALG{DMBC} and
\ALG{DMBC$^+$}) will presented in Section~\ref{sec:search-algorithm} and
Section~\ref{sec:search-impr-algorithm}. In order to present those improvements
uniformly and concisely, we generalize and modify some previously existing
concepts in the next four subsections.

\subsection{Domain Reduction Operators}
\label{sec:search-dr-operators}%

First, we define the concept of a domain reduction operator as follows.

\begin{definition}[Domain Reduction Operator, $\DR$]
\label{def:dr-operator}%
Given a sequence $X$ of $n$ real variables associated with domains $\cl{D}$. A
\emph{domain reduction operator} $\DR$ for numerical constraints is a function
that takes as input a box $\IA{x} \in \II^n$ contained in $\cl{D}$ and a finite
set $\cl{C}$ of constraints on $X$, and that returns a box in $\II^n$, denoted
by $\DR(\IA{x}, \cl{C})$, satisfying the following properties:
\begin{align}
    & \tn{(Contractiveness)} & \DR(\IA{x}, \cl{C}) & \subseteq \IA{x},
    \\
    & \tn{(Correctness)} & \DR(\IA{x}, \cl{C}) & \supseteq \IA{x} \cap
    \bigcap_{C \in \cl{C}} C.
\end{align}
\end{definition}

\begin{figure}[!htb]
\vspace{-4mm}
    \centerline{\includegraphics[width=9.5cm]{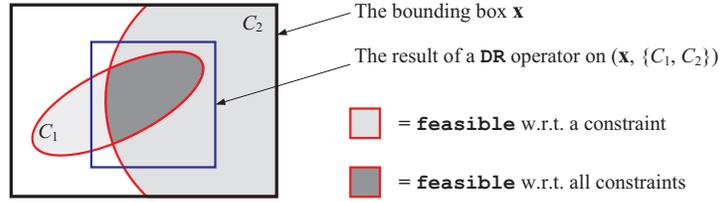}}
\caption[An example of a domain reduction (\DR) operator]{A domain reduction
(\DR) operator is applied to a box $\IA{x}$ and a constraint set $\set{C_1,
C_2}$.}
\label{fig:dr-operator}%
\end{figure}

A domain reduction operator has also been referred to as a narrowing operator,
contracting operator, or contractor in literature. We adopt the terminology
\emph{domain reduction operator}, because it is mnemonic and the terminology
\emph{domain reduction} has been widely accepted in many fields, not only in
constraint programming.

\begin{definition}[Monotonicity]
\label{def:monotonicity}%
Given a sequence $X$ of $n$ real variables associated with domains $\cl{D}$. A
domain reduction operator $\mu$ is said to be \emph{monotonic} if, for every
set \cl{C} of constraints on $X$, we have
\begin{equation}
    \forall \IA{x}, \IA{x}' \in \II^n,\ \ \IA{x} \subseteq \cl{D},\ \ \IA{x}'
    \subseteq \cl{D}\;\; : \;\; \IA{x} \subseteq \IA{x}'\ \ \OIF\ \ \mu(\IA{x},
    \cl{C}) \subseteq \mu(\IA{x}', \cl{C}).
\end{equation}
\end{definition}

In constraint programming, domain reduction operators are usually constructed
by enforcing either box consistency \cite{BenhamouF-MA-VH:1994}, hull
consistency \cite{BenhamouF-O:1992,BenhamouF-O:1997}, or $k\BB$-consistency
\cite{LhommeO:1993}. Although these domain reduction operators enjoy the
monotonicity, many domain reduction operators do not enjoy the monotonicity but
are still very efficient in practice. The concept of a domain reduction
operator is depicted in Figure~\ref{fig:dr-operator}. Other examples of domain
reduction operators are depicted in Figure~\ref{fig:operators-ex}. The
following property is straightforward but interesting for constraint solving.

\begin{theorem}
\label{thm:oc-operator-empty}%
Given a set $\cl{C}$ of constraints on a sequence of $n$ real variables
associated with domains $\cl{D}$. Suppose $\IA{x} \in \II^{n}$ is a box
contained in $\cl{D}$. If there exists a domain reduction operator $\DR$ that
maps $(\IA{x}, \cl{C})$ to an empty set (i.e., $\DR(\IA{x}, \cl{C}) =
\emptyset$), then $\cl{C}$ is inconsistent in $\IA{x}$; that is,
\begin{equation}
    \IA{x} \subseteq \lnot \cl{C} \qquad (\textrm{where } \lnot \cl{C} \equiv
    \cl{D} \setminus \bigcap_{C \in \cl{C}} C).
\end{equation}
\end{theorem}

\begin{proof}
It follows from the correctness of domain reduction operators
(Definition~\ref{def:dr-operator}) that $\IA{x} \cap \bigcap_{C \in \cl{C}} C =
\emptyset$. Thus, we have $\IA{x} \subseteq \lnot \cl{C}$.
\end{proof}

\subsection{Complementary Boxing Operators}
\label{sec:search-cb-operators}%

We recall that in \cite{BenhamouF-G:2000,BenhamouF-G-L-C:2004}, a negation test
takes as input a universally quantified constraint of the form $\forall t \in
D_t: f(x, t) \le 0$ and performs a kind of domain reduction operator on its
negation, $f(x, t) > 0$. It encloses all possibly infeasible regions and the
remaining is feasible.

The negation test has been extended to solve classic numerical constraints by
\citeN{SilaghiMC-SH-F:2001}, in which a negation test is applied to
inequalities of the form $f(x) \le 0$. The search algorithm in
\cite{SilaghiMC-SH-F:2001}, called \ALG{UCA6}, employs the negation test to
enclose the negations of all individual constraints and then chooses the best
result to guide the domain splitting during search. A concise description of
the \ALG{UCA6} algorithm is presented in
Section~\ref{sec:new-search-algorithm}.

For convenience, we define a kind of operator to generalize the idea of a
negation test, and then give several interesting properties. The generalized
operator is called the \emph{complementary boxing operator}, and the
corresponding splitting operator is called the \emph{box splitting operator}.

\begin{definition}[Complementary Boxing Operator, $\CB$]
\label{def:cbc-operator}%
Given a sequence $X$ of $n$ real variables associated with domains $\cl{D}$. A
\emph{complementary boxing operator} is a function $\CB$ that takes as input a
box $\IA{x} \in \II^n$ contained in $\cl{D}$ and a finite set $\cl{C}$ of
constraints on $X$, and that returns a box in $\II^n$, denoted by $\CB(\IA{x},
\cl{C})$, satisfying the following properties:
\begin{align}
    & \tn{(Contractiveness)} & \CB(\IA{x}, \cl{C}) & \subseteq \IA{x},
    \\
    & \tn{(Complementariness)} & \IA{x} \setminus \CB(\IA{x}, \cl{C}) &
    \subseteq \bigcap_{C \in \cl{C}} C.
\end{align}
\end{definition}

\begin{figure}[!htb]
\vspace{-2em}
    \centerline{\includegraphics[width=9.5cm]{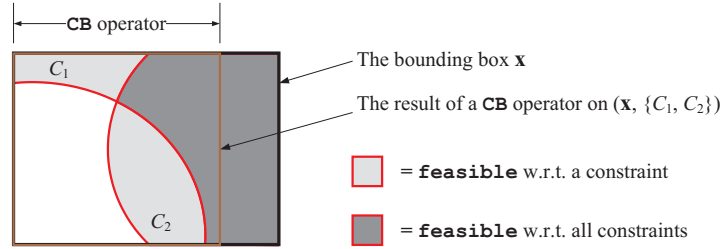}}
\caption[An example of a complementary boxing (\CB) operator]{An example of a
complementary boxing (\CB) operator applied to a box $\IA{x}$ and a set
$\set{C_1, C_2}$ of two constraints.}
\label{fig:cb-operator}%
\end{figure}

\begin{figure}[!htb]
    \centerline{\includegraphics[width=9.6cm]{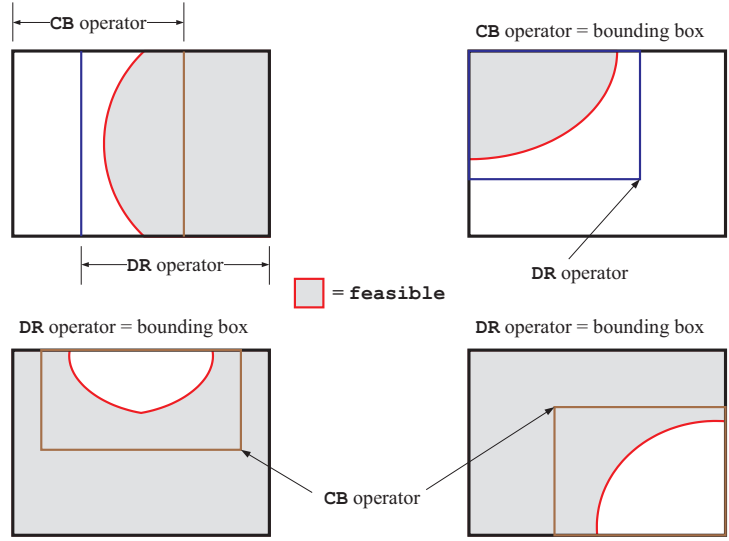}}
\caption[Domain reduction (\DR) operators and complementary boxing (\CB)
operators]{Examples of domain reduction (\DR) operators and complementary
boxing (\CB) operators.}
\label{fig:operators-ex}%
\end{figure}

A box resulting from the application of a complementary boxing operator to a
bounding box $\IA{x}$ and a set $\cl{C}$ of constraints is called a
\emph{complementary box} of $\cl{C}$ within $\IA{x}$. The term
\emph{complementary boxing} refers to the process of computing a complementary
box. The concept of a complementary boxing operator is depicted in
Figure~\ref{fig:cb-operator}. Additionally, Figure~\ref{fig:operators-ex}
illustrates the outcomes of domain reduction operators and complementary boxing
operators when applied to the same bounding boxes, in some typical situations.

The complementariness of complementary boxing operators means that the
complementary boxing allows isolating certain regions, namely $\IA{x} \setminus
\CB(\IA{x}, \cl{C})$, of which the points entirely satisfy all the constraints
in $\cl{C}$. Especially, if the application of a complementary boxing operator
to a box and a constraint results in an empty set, then the box completely
satisfies that constraint. Similarly, if the application of a complementary
boxing operator to a box with the whole set of constraints results in an empty
set, then the box is completely feasible. The following theorem states this
property formally.

\begin{theorem}
\label{thm:cbc-operator-empty}%
Given a set $\cl{C}$ of constraints on a sequence of $n$ real variables
associated with domains $\cl{D}$. Suppose $\IA{x}$ is a box contained in
$\cl{D}$. If there exists a complementary boxing operator $\CB$ that maps
$(\IA{x}, \cl{C})$ to an empty set (i.e., $\CB(\IA{x}, \cl{C}) = \emptyset$),
then $\cl{C}$ is satisfied with every point in $\IA{x}$; that is, $\IA{x}
\subseteq \bigcap_{C \in \cl{C}} C$.
\end{theorem}

A complementary boxing operator can be constructed from a domain reduction
operator as stated in the following theorem.

\begin{theorem}
\label{thm:dr-to-cb-operator}%
Given a domain reduction operator $\DR$. The function $f$ defined as $f(\IA{x},
\cl{C}) \equiv \DR(\IA{x}, \lnot \cl{C})$ is a complementary boxing operator.
\end{theorem}

\begin{proof}
By definition $\IA{x}_f = f(\IA{x}, \cl{C}) = \DR(\IA{x}, \lnot\cl{C})$. The
contractiveness of domain reduction operators implies that $\IA{x}_f \subseteq
\IA{x}$. That is, $f$ enjoys the contractiveness of complementary boxing
operators. In addition to that, the correctness of domain reduction operators
implies that
\begin{equation}
\label{eqn:dr-cb-proof-1}%
    \IA{x} \cap \lnot\cl{C} \subseteq \IA{x}_f
\end{equation}
It follows from \eqref{eqn:dr-cb-proof-1} that, for all $x \in \IA{x} \setminus
\IA{x}_f$, we have $x \notin \IA{x} \cap \lnot\cl{C}$; thus, $x \notin
\lnot\cl{C} \equiv \cl{D} \setminus \bigcap_{C \in \cl{C}} C$, and $x \in
\bigcap_{C \in \cl{C}} C$ because $x \in \IA{x} \subseteq \cl{D}$. That is, we
have $\IA{x} \setminus \IA{x}_f \subseteq \bigcap_{C \in \cl{C}} C$. Thus, $f$
enjoys the complementariness of complementary boxing operators.
\end{proof}

\begin{theorem}
\label{thm:cb-to-dr-operator}%
Given a complementary boxing operator $\CB$. The function $f$ defined as
$f(\IA{x}, \cl{C}) \equiv \CB(\IA{x}, \lnot \cl{C})$ is a domain reduction
operator.
\end{theorem}

\begin{proof}
By definition $\IA{x}_f = f(\IA{x}, \cl{C}) = \CB(\IA{x}, \lnot\cl{C})$. The
contractiveness of complementary boxing operators implies that $\IA{x}_f
\subseteq \IA{x}$; that is, $f$ enjoys the contractiveness of domain reduction
operators. In addition to that, the complementariness of complementary boxing
operators implies that
\begin{eqnarray}
\label{eqn:cb-dr-proof-1}%
    && \IA{x} \setminus \IA{x}_f \subseteq \lnot \cl{C} = \cl{D} \setminus
    \bigcap_{C \in \cl{C}} C
\end{eqnarray}
It follows from \eqref{eqn:cb-dr-proof-1} that, for all $x \in \IA{x}_f$, we
have $x \notin \IA{x} \cap \lnot\cl{C}$; thus, $x \notin \cl{D} \setminus
\bigcap_{C \in \cl{C}} C$ and $x \in \bigcap_{C \in \cl{C}} C$ because $x \in
\IA{x}_f \subseteq \cl{D}$. That is, we have $\IA{x} \setminus \IA{x}_f
\subseteq \cl{C}$. This means that $f$ enjoys the complementariness of
complementary boxing operators.
\end{proof}

It follows from Theorem~\ref{thm:dr-to-cb-operator} and
Theorem~\ref{thm:cb-to-dr-operator} that complementary boxing operators can be
constructed from domain reduction operators and vice versa. In other words,
they are dual to each other. In particular, let $\cl{C} = \set{C_1, \dots,
C_k}$ be a set of $k$ constraints. A complementary boxing operator can be
constructed by $\CB(\IA{x}, \cl{C}) := \DR(\IA{x}, \lnot \cl{C}) = \DR(\IA{x},
C_0)$, where $C_0$ is the disjunction of constraints $\lnot C_1, \dots, \lnot
C_k$. In a system that does not accept disjunctive constraints, we can relax
them by taking the (interval) union of complementary boxes, as stated in
following theorem.

\begin{theorem}
\label{thm:cb-relax}%
Consider a sequence $X$ of $n$ real variables associated with domains $\cl{D}$.
Let $\cl{C} = \set{C_1, \dots, C_k}$ be a set of constraints on $X$ and
$\set{\DR_1, \dots, \DR_k}$ a set of domain reduction operators for $X$.
Suppose $\set{C_1', \dots, C_k'}$ is a set of constraints on $X$ such that
$\lnot C_i \equiv \cl{D} \setminus C_i \subseteq C_i'$ for all $i = 1, \dots,
k$. Then the operator defined by the following rule is a complementary boxing
operator:
\vspace{-0.3em}%
\begin{equation}
\label{eqn:cb-by-union-boxes}
    \forall \IA{x} \in \II^n, \IA{x} \subseteq \cl{D} : f(\IA{x}, \cl{C})
    \equiv \Hull{\bigcup_{i = 1}^{k} \DR_i(\IA{x}, C_i')},
\end{equation}
\end{theorem}

\begin{proof}
The contractiveness of $f$ is obvious because $\DR_i(\IA{x}, C_k') \subseteq
\IA{x}$. We now prove the complementariness. For every $x \in \IA{x} \setminus
\bigcup_{i = 1}^{k} \DR_i(\IA{x}, C_i')$ and $i \in \set{1, \dots, k}$:
\begin{align*}
    &\ x \notin \DR_i(\IA{x}, C_i') \supseteq \IA{x} \cap C_i'
    \\
    \OIF &\ x \notin \IA{x} \cap C_i'
    \\
    \OIF &\ x \notin C_i' \supseteq \lnot C_i
    \\
    \OIF &\ x \notin \cl{D} \setminus C_i
    \\
    \OIF &\ x \in C_i & (\tn{since } x \in \cl{D}).
\end{align*}
Thus, $x \in \bigcap_{i=1}^{k} C_i$. Therefore, we have
\vspace{-0.3em}%
\[
    \IA{x} \setminus f(\IA{x}, \cl{C}) \subseteq \IA{x} \setminus \bigcup_{i =
    1}^{k} \DR_i(\IA{x}, C_i') \subseteq \bigcap_{i=1}^{k} C_i.
\]
This is the complementariness as required.
\end{proof}

The negation $\lnot C$ of a numerical constraint $C$ of the form $f(x) \diamond
0$ (where $\diamond$ is either $\le$, $<$, $\ge$, $>$, $=$, or $\ne$) is the
constraint $f(x)\ \widetilde{\diamond}\ 0$ (where $\widetilde{\diamond}$ is
either $>$, $\ge$, $<$, $\le$, $\ne$, or $=$, respectively). In practice, some
implementations of domain reduction operators only accept constraints that are
defined with the relations $\le$ and $\ge$, but not with the relations $<$ and
$>$. For example, a constraint $C_i$ of the form $C_i \equiv (f(x) \le 0)$ has
the negation of the form $\lnot C_i \equiv (f(x) > 0)$, which is not accepted
in some implementations. Fortunately, we can safely use $C_i' \equiv (f(x) \ge
0)$ in the complementary boxing operator defined by
\eqref{eqn:cb-by-union-boxes} because $\lnot C_i \subseteq C_i'$ holds.

The monotonicity of complementary boxing operators is defined similarly to that
of domain reduction operators (see Definition~\ref{def:monotonicity}). The
following theorem gives a way to construct a (monotonic) inclusion test.

\begin{theorem}
\label{thm:fc-checker}%
Let $X$ be a sequence of $n$ real variables, $\set{\DR_1, \dots, \DR_n}$ a set
of (respectively, monotonic) domain reduction operators and $\set{\CB_1, \dots,
\CB_n}$ a set of (respectively, monotonic) complementary boxing operators,
where $\DR_k$ and $\CB_k$ ($k = 1, \dots, n$) are defined in $k$ dimensions.
Let $\FC$ be a function that takes as input a box $\IA{x} \in \II^n$ and a
finite set $\cl{C}$ of constraints on a subsequence $Y$ of size $k$ of $X$, and
that returns either $\CONST{feasible}$, $\CONST{infeasible}$, or
$\CONST{unknown}$ such that:
\begin{align}
    \FC(\IA{x}, \cl{C}) & = \CONST{infeasible} & \IFF &&
    \DR_k(\proj{\IA{x}}{Y}, \cl{C}) & = \emptyset;
    \label{eqn:fc-eq1}
    \\
    \FC(\IA{x}, \cl{C}) & = \CONST{feasible} & \IFF &&
    \CB_k(\proj{\IA{x}}{Y}, \cl{C}) & = \emptyset.
    \label{eqn:fc-eq2}
\end{align}
Then $\FC$ is a (respectively, monotonic) inclusion test. If we restrict the
codomain of $\FC$ to either $\set{\CONST{infeasible}, \CONST{unknown}}$ or
$\set{\CONST{feasible}, \CONST{unknown}}$, then the result still holds if we
use either \eqref{eqn:fc-eq1} or \eqref{eqn:fc-eq2}, respectively, to construct
$\FC$.
\end{theorem}

\begin{proof}
It follows directly from Definition~\ref{def:inclusion-test},
Definition~\ref{def:fc-checker}, Definition~\ref{def:monotonicity},
Theorem~\ref{thm:oc-operator-empty} and Theorem~\ref{thm:cbc-operator-empty}.
\end{proof}

\subsection{Domain Splitting Operators}
\label{sec:search-splitting-opt}%

First, we recall the concept of a bisection, where an interval (i.e., a side)
of a box is dichotomized into two parts.

\begin{definition}[Dichotomous Splitting Operator, \DS]
\label{def:ds-operator}%
A \emph{dichotomous splitting} (\DS) operator is a function $\DS: \II^n \to
2^{\II^n}$ that takes as input a box in $\II^n$, and that returns two
(disjoint) boxes in $\II^n$ resulting from splitting a side of the input box
into two halves.
\end{definition}

\begin{example}
Consider a box $\IA{x} \equiv (\IA{x}_1, \dots, \IA{x}_n)^\TT \in \II^n$, where
$\IA{x}_i = \rintv{\Lb{x}_i}{\Ub{x}_i}$. A dichotomous splitting operator $\DS$
splits $\IA{x}$ into two disjoint boxes: $\IA{x}' = (\IA{x}_1, \dots,
\IA{x}_{i-1}, \IA{x}_i', \IA{x}_{i+1}, \dots, \IA{x}_n)^\TT$ and $\IA{x}'' =
(\IA{x}_1, \dots, \IA{x}_{i-1}, \IA{x}_i'', \IA{x}_{i+1}, \dots,
\IA{x}_n)^\TT$, where $\IA{x}_i' = \rintv{\Lb{x}_i}{\Mid(\IA{x})}$ and
$\IA{x}_i'' = \rintv{\Mid(\IA{x})}{\Ub{x}_i}$. Note that $\IA{x}_i' \cap
\IA{x}_i'' = \emptyset$; thus, $\IA{x}' \cap \IA{x}'' = \emptyset$.
\end{example}

Second, we define the concept of a box splitting operator, which splits around
a complementary box in order to isolate \emph{feasible regions} w.r.t. a subset
of constraints.

\begin{definition}[Box Splitting Operator, \BS]
\label{def:bs-operator}%
A \emph{box splitting} (\BS) operator is a function $\BS: \II^n \times \II^n
\to 2^{\II^n}$ which takes as input two boxes such that the former contains the
latter, and which sequentially splits the outer box along some facets of the
inner one. The outcome is a sequence of disjoint boxes.
\end{definition}

\begin{figure}[!htb]
    \centerline{\includegraphics[width=10cm]{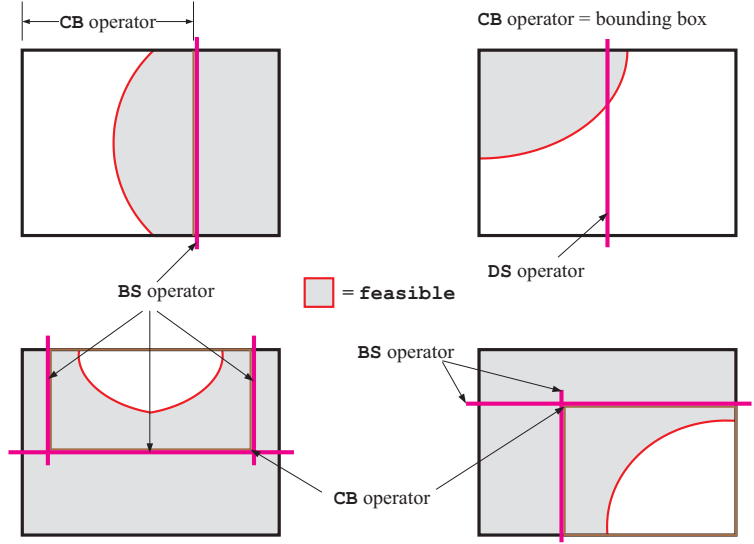}}
\caption[Examples of box splitting (\BS) and dichotomous splitting (\DS)
operators]{Examples of box splitting (\BS) and dichotomous splitting (\DS)
operators. In box splitting, all boxes excepted the complementary box are
feasible w.r.t. the considered constraints.}
\label{fig:bs-operator}%
\end{figure}

In fact, the concept of a box splitting operator is a slight generalization of
the splitting operator proposed in \cite{VanIwaardenRJ:Thesis:1996}. The
original splitting operator gives a way to split a region surrounding a box,
provided that this box contains at most one optimal solution to a considered
optimization problem.

In our algorithm, a box splitting operator that takes as input a domain box and
a complementary box resulting from the application of a complementary boxing
operator is applied in combination with a dichotomous splitting operator. The
dichotomous splitting operator is used when either the complementary boxing
operator produces no reduction or the box splitting operator results in too
small boxes. Figure~\ref{fig:bs-operator} illustrates the concept of a box
splitting operator for this purpose.

\section{Basic Branch-and-Prune Search Algorithms}
\label{sec:search-algorithm}%

In this section, we first present a generic search in the branch-and-prune
framework and then present three search algorithms for NCSPs with emphasis on
problems with continuums of solutions.

\subsection{A Generic Branch-and-Prune Search Algorithm}
\label{sec:bnp-search-algorithm}%

We now present a generic search technique, called \ALG{BnPSearch}, for solving
NCSPs in the branch-and-prune framework, the most common framework for the
complete solution of NCSPs. The main steps of this generic search technique are
described in Algorithm~\ref{alg:bnp-search}. Although this technique is not the
most general one, it is still capable of generalizing the majority of the
existing branch-and-prune techniques.

\begin{algorithm}[!b]
\caption{\ALG{BnPSearch} -- a generic branch-and-prune search algorithm}%
\label{alg:bnp-search}%
    \SetVline
    \KwIn{a CSP $\cl{P} \equiv (\cl{V}_0, \cl{D}_0, \cl{C}_0)$.}
    \KwOut{$\cl{L}_{\forall}$, $\cl{L}_{\varepsilon}$.
    \cright{Lists of boxes in inner and boundary union approximations, respectively.}}

    \lIf{\ref{alg:prunecheck}($\cl{V}_0$, $\cl{D}_0$, $\cl{C}_0$, $\varepsilon$, $\FC$, \VAR{WaitList})}
        \Return;
    \cright{On page~\pageref{alg:prunecheck}.}

    \While{$\VAR{WaitList} \neq \emptyset$}
    {
        Get a couple $(\cl{D}, \cl{C})$ from \VAR{WaitList};
        \cnext{$\forall i \in \{1, \dots, k\}$: $\cl{D}_i \subseteq \cl{D}$,
        $\cl{C}_i \subseteq \cl{C}$.}

        Split the CSP $(\cl{V}_0, \cl{D}, \cl{C})$ into sub-CSPs
        $\set{(\cl{V}_0, \cl{D}_1, \cl{C}_1), \dots, (\cl{V}_0, \cl{D}_k, \cl{C}_k)}$;

        \For{$i := 1, \dots, k$}
        {
            \vspace{-1em}
            \cright{Do branching.}

            \If{$\cl{C}_i = \emptyset$}
            {
                $\cl{L}_{\forall} := \cl{L}_{\forall} \cup \set{\cl{D}_i}$;
                \cright{All points in $\cl{D}_i$ are solutions.}
                \CMD{continue for};
            }
            \ref{alg:prunecheck}($\cl{V}_0$, $\cl{D}_i$, $\cl{C}_i$, $\varepsilon$, $\FC$, \VAR{WaitList});
            \cright{On page~\pageref{alg:prunecheck}.}
        }
    }
\end{algorithm}

\begin{function}[!b]
\caption{\ALG{PruneCheck}($\cl{V}$, $\cl{D}$, $\cl{C}$, $\varepsilon$, $\FC$, \VAR{WaitList})}%
\label{alg:prunecheck}%
    \SetVline
    $\cl{D} := \ALG{Prune}(\cl{V}, \cl{D}, \cl{C})$;
    \cright{Prune the domains in $\cl{D}$ by using domain reduction techniques.}

    \lIf{$\cl{D} = \emptyset$}
        \Return{$\CONST{true}$};

    \If{all the domains in $\cl{D}$ are not amenable to be split (w.r.t.
    $\varepsilon$)}
    {
        $\cl{L}_{\varepsilon} := \cl{L}_{\varepsilon} \cup \set{(\cl{D}, \cl{C})}$;
        \Return{$\CONST{true}$};
    }

    \PRO{put}($\VAR{WaitList} \leftarrow (\cl{D}, \cl{C})$)\;
    \Return{\CONST{false}}; \cright{The problem has not been solved yet.}
\end{function}

Taking a CSP as input, the \ALG{BnPSearch} algorithm produces two lists:
$\cl{L}_{\forall}$ and $\cl{L}_{\varepsilon}$. The first list,
$\cl{L}_{\forall}$, is an inner approximation of the solution set. The second
list, $\cl{L}_{\varepsilon}$, consists of couples, each consists of a sequence
of domains and a set of running constraints in these domains. When removing the
running constraints, $\cl{L}_{\varepsilon}$ will become a boundary
approximation of the solution set in association with the inner approximation
$\cl{L}_{\forall}$. Each couple in $\cl{L}_{\varepsilon}$ constitues a CSP
which will need to be explored further when reducing the value of the precision
$\varepsilon$ and continuing the solving process. For simplicity, we may omit
the running constraints in $\cl{L}_{\varepsilon}$. The pruning procedure
\ALG{Prune} can be any combination of domain reduction techniques, which is not
necessarily fixed for all steps of the algorithm.

\begin{notation}
The notations used in the algorithms in the rest of the paper follow the
following conventions:
\begin{itemize}
    \item The notations $\mb{B}$, $\mb{B}_i$ and $\mb{B}'$ denote
    relevant bounding boxes in $\II^n$; that is, the vectors of domains of the
    considered NCSPs.

    \item The notations $\cl{C}$, $\cl{C}_i$, $\cl{C}'$ and $\cl{C}''$
    denote relevant sets of constraints.

    \item The notation $\mb{CB}_c$ denotes a complementary box w.r.t. a
    constraint, $c$.

    \item The notation $\UNION{I}$ and $\UNION{B}$ denote the global lists that
    accumulate computed boxes of inner and boundary union approximations,
    respectively.

    \item The notations $\DR$, $\CB$, and $\FC$ denote some domain reduction
    operator, complementary boxing operator, and inclusion test,
    respectively.
\end{itemize}
\end{notation}

\subsection{Classic Branch-and-Prune Search Algorithms}
\label{sec:classic-search-algorithm}

As mentioned in Section~\ref{sec:overview-bnp}, most complete methods for
solving NCSPs integrate domain reduction techniques, existence tests,
uniqueness tests, exclusion tests, or inclusion tests into a \emph{bisection}
search strategy. The most successful solution methods enhance this process by
applying domain reduction techniques such as \emph{consistency} techniques to
the constraint system after each split. This policy is referred to as
\emph{dichotomous maintaining bounds by consistency} (\ALG{DMBC}). A generic
\ALG{DMBC} algorithm is presented in Algorithm~\ref{alg:dmbc}, where
$\varepsilon$ is a positive precision (vector) and $\FC$ is a monotonic
inclusion test (see Definition~\ref{def:fc-checker}).

In most of the known \ALG{DMBC} techniques, the search strategy is to perform
splitting intervals until the intervals are canonical or their widths are not
greater than a predetermined precision $\varepsilon$. That is, these techniques
are able to achieve a predetermined precision $\varepsilon$
(Definition~\ref{def:interval-precision}). In general, the \ALG{DMBC} algorithm
cannot detect feasible boxes, thus mainly addressing NCSPs with isolated
solutions. When solving NCSPs with continuums of solutions, we can replace
Function \ref{alg:prunecheckDMBC} of the \ALG{DMBC} algorithm by Function
\ref{alg:prunecheckDMBCp} (on page \pageref{alg:prunecheckDMBCp}). The obtained
algorithm is thus called \ALG{DMBC$^+$}.

\begin{algorithm}[!tb]
\caption{\ALG{DMBC} -- an instance of the \ALG{BnPSearch} algorithm}%
\label{alg:dmbc}%
    \SetVline
    \KwIn{a bounding box $\mb{B}_{0}$, a constraint set $\cl{C}_{0}$,
    $\varepsilon \in \Rp^n$, a monotonic inclusion test $\FC$.}

    \KwOut{an inner union approximation $\UNION{I}$, a boundary union
    approximation $\UNION{B}$.}

    \lIf{\ref{alg:prunecheckDMBC}($\mb{B}_0$, $\cl{C}_0$, $\varepsilon$, $\FC$,
    \VAR{WaitList})}
    {
        \Return;
    }
    \cright{Page~\pageref{alg:prunecheckDMBC}.}

    \While{$\VAR{WaitList} \neq \emptyset$}
    {
        Get a box $\mb{B}$ from \VAR{WaitList};

        $(\mb{B}_1, \mb{B}_2) := \PRO{Bisect}(\mb{B}$);
        \cright{$\forall i \in \set{1, 2}$: $\mb{B}_i \subseteq \mb{B}$.}

        \ref{alg:prunecheckDMBC}($\mb{B}_1$, $\cl{C}_0$, $\varepsilon$, $\FC$,
        \VAR{WaitList});
        \cright{On page~\pageref{alg:prunecheckDMBC}.}
        \ref{alg:prunecheckDMBC}($\mb{B}_2$, $\cl{C}_0$, $\varepsilon$, $\FC$,
        \VAR{WaitList});
        \cright{On page~\pageref{alg:prunecheckDMBC}.}
    }
\end{algorithm}

\begin{function}[!tb]
\caption{\ALG{PruneCheckDMBC}($\mb{B}$, $\cl{C}$, $\varepsilon$, $\FC$, \VAR{WaitList})}%
\label{alg:prunecheckDMBC}%
    \SetVline

    $\mb{B}' := \DR(\mb{B}, \cl{C})$; \cright{Reduce domains.}

    \lIf{$\mb{B}' = \emptyset$} \Return{\CONST{true}};
    \cright{$\mb{B}$ is infeasible, the problem has been solved.}

    \If{$\mb{B}$ is canonical or $\Wid(\mb{B}) \le \varepsilon$}
    {
        \ref{alg:checkEpsilon}($\mb{B}'$, $\cl{C}$, $\FC$);
        \cright{On page~\pageref{alg:checkEpsilon}.}
        \Return{\CONST{true}};
    }

    $\PRO{put}(\VAR{WaitList} \leftarrow \mb{B}')$;
    \cright{Put the current problem into the waiting list.}

    \Return{\CONST{false}}; \cright{The problem has not been solved yet.}
\end{function}

\begin{function}[!tb]
\caption{\ALG{CheckEpsilon}($\mb{B}$, $\cl{C}$, $\FC$)}%
\label{alg:checkEpsilon}%
    \SetVline
    \uIf{$(\VAR{Result} := \FC(\mb{B}, \cl{C})) = \CONST{feasible}$}
    {
        \vspace{-1em}
        \cright{Identify the feasibility of $\mb{B}$ w.r.t. $\cl{C}$.}
        $\UNION{I} :=  \UNION{I} \cup \set{\mb{B}}$;
        \cright{$\mb{B}$ is feasible, store it into the list of inner boxes.}
    }
    \ElseIf{$\VAR{Result} = \CONST{unknown}$}
    {
        $\UNION{B} := \UNION{B} \cup \set{\mb{B}}$;
        \cright{$\mb{B}$ is undiscernible, store it into the list of boundary
        boxes.}
    }
\end{function}

\begin{function}[!tb]
\caption{\ALG{PruneCheckDMBC$^+$}($\mb{B}$, $\cl{C}$, $\varepsilon$, $\FC$, \VAR{WaitList})}%
\label{alg:prunecheckDMBCp}%
    \SetVline

    $\mb{B}' := \DR(\mb{B}, \cl{C})$; \cright{Reduce domains.}

    \lIf{$\mb{B}' = \emptyset$} \Return{\CONST{true}};
    \cright{$\mb{B}$ is infeasible, the problem has been solved.}

    \If{$\mb{B}$ is canonical \CMD{or} $\Wid(\mb{B}) \le \varepsilon$}
    {
        \ref{alg:checkEpsilon}($\mb{B}'$, $\cl{C}$, $\FC$);
        \cright{On page~\pageref{alg:checkEpsilon}.}
        \Return{\CONST{true}};
    }

    \lnl{lnl:dmbcp-diff}
    \If{$\FC'(\mb{B}', \cl{C}) = \CONST{feasible}$}
    {
        \vspace{-1em}
        \cright{This can be optionally replaced with the check $\CB(\mb{B}', \cl{C}) = \emptyset$.}

        $\UNION{I} :=  \UNION{I} \cup \set{\mb{B}'}$;
        \cright{$\mb{B}$ is feasible, store it into the list of inner boxes.}
        \Return{\CONST{true}};
    }

    $\PRO{put}(\VAR{WaitList} \leftarrow \mb{B}')$;
    \cright{Put the current problem into the waiting list.}

    \Return{\CONST{false}}; \cright{The problem has not been solved yet.}
\end{function}

The difference (at Line~\ref{lnl:dmbcp-diff}) between Function
\ref{alg:prunecheckDMBC} and Function \ref{alg:prunecheckDMBCp} is that the
latter resorts to an inclusion test, $\FC'$ (not $\FC$), to check if a box is
feasible (in other words, it is an inner box). This can also be replaced with a
complementary boxing operator, $\CB$, checking if it returns an empty set.
Hence, the \ALG{DMBC$^+$} algorithm is able to detect feasible boxes, provided
that the inclusion test $\FC'$ (or the complementary boxing operator $\CB$) is
sufficiently efficient. If the inclusion test $\FC'$ is implemented using an
interval form of functions as in \cite{JaulinL-W:1993,JaulinL-K-D-W:Book:2001},
then the \ALG{DMBC$^+$} will become the \ALG{SIVIA}
algorithm\footnote{\ALG{SIVIA} is the abbreviation of \emph{set inverter via
interval analysis}.} in \cite{JaulinL-W:1993,JaulinL-K-D-W:Book:2001}.

Because of the finite nature of floating-point numbers on computers, it is easy
to prove that both the \ALG{DMBC} and \ALG{DMBC$^+$} algorithms terminate
(i.e., the waiting list $\VAR{WaitList}$ becomes empty) after a finite number
of steps and are of the precision $\varepsilon$ w.r.t. the monotonic inclusion
test $\FC$ (see Definition~\ref{def:interval-precision}).

\subsection{New Branch-and-Prune Search Algorithms}
\label{sec:new-search-algorithm}

The \ALG{DMBC} and \ALG{DMBC$^+$} algorithms often generate verbose inner and
outer union approximations, especially when solving NCSPs with continuums of
solutions. The first reason is that entirely feasible boxes may be
unnecessarily split. The second reason is that all constraints are always
considered in computations, even when some of them are completely satisfied by
all points in considered domains.

Before presenting new search algorithms, we define the following terms.

\begin{definition}
\label{def:active-var}%
Given a sequence $X$ of $n$ real variables $(x_1, \dots, x_n)$, and a vector
$\varepsilon = (\varepsilon_1, \dots, \varepsilon_n)^\TT \in \Rp^n$. Consider a
box $\IA{x} = (\IA{x}_1, \dots, \IA{x}_n)^\TT \in \II^n$ and a finite set
$\cl{C}$ of constraints on subsequences of $X$. A variable $x_i$ is called an
\emph{active variable in $\IA{x}$ w.r.t. $\cl{C}$ and $\varepsilon$} if it
occurs in at least one constraint in $\cl{C}$, $\Wid(\IA{x}_i) \equiv
\sup(\IA{x}_i) - \inf(\IA{x}_i) > \varepsilon_i$ holds and $\IA{x}_i$ is not
canonical. A variable $x_i$ is called an \emph{inactive variable in $\IA{x}$
w.r.t. $\cl{C}$ and $\varepsilon$} if it is not active in $\IA{x}$ w.r.t.
$\cl{C}$ and $\varepsilon$.
\end{definition}

Inspired by the idea of the \ALG{ICAb5} algorithm \cite{BenhamouF-G:2000}, we
present in Algorithm~\ref{alg:uca5} (on page~\pageref{alg:uca5}) a simplified
version, called \ALG{UCA5}, of the \ALG{UCA6} algorithm
\cite{SilaghiMC-SH-F:2001}, where $\ALG{UCA}$ stands for
\emph{union-constructing approximation}. The \ALG{UCA5} algorithm takes as
input an NCSP $\cl{P} \equiv (\cl{V}, \cl{C}_0, \mb{B}_0)$ and returns an inner
union approximation $\UNION{I}$ and a boundary union approximation $\UNION{B}$
of the solution set of $\cl{P}$. The outer union approximation of the solution
set can be computed by $\UNION{O} := \UNION{I} \cup \UNION{B}$. Roughly
speaking, the \ALG{UCA5} algorithm proceeds by recursively repeating three main
steps/processes:

\begin{enumerate}
    \item\label{stp:uca5-1} Use domain reduction ($\DR$) operators to reduce
    the current bounding box, which plays the role of domains, to a
    narrower one (see Line~\ref{lnl:uca5-prune} in Algorithm~\ref{alg:uca5} and
    Line~\ref{lnl:uca5-dr} in Function \ref{alg:prunecheck-uca5}).

    \item\label{stp:uca5-2} Use complementary boxing ($\CB$) operators to
    search for a complementary box w.r.t. some running constraint and the new
    bounding box obtained at Step~\ref{stp:uca5-1} (Line~\ref{lnl:uca5-cb} in
    Function \ref{alg:branch-uca5}). During this search, the constraints that make
    empty complementary boxes are removed (Line~\ref{lnl:uca5-c-empty} in
    Function \ref{alg:branch-uca5}).

    \item\label{stp:uca5-3} Combine dichotomous splitting ($\DS$) operators
    with box splitting ($\BS$) operators to split the current problem into
    subproblems (see Line~\ref{lnl:uca5-split} in Algorithm~\ref{alg:uca5} and
    Line~\ref{lnl:uca5-bs} to Line~\ref{lnl:uca5-ds} in
    Function \ref{alg:branch-uca5}).
\end{enumerate}

\begin{algorithm}[!tb]
\caption{\ALG{UCA5} -- a new branch-and-prune search algorithm}%
\label{alg:uca5}%
    \SetVline
    \KwIn{a bounding box $\mb{B}_{0}$, a constraint set $\cl{C}_{0}$,
    $\varepsilon \in \Rp^n$, a monotonic inclusion test $\FC$.}

    \KwOut{an inner union approximation $\UNION{I}$, a boundary union
    approximation $\UNION{B}$.}

    $\UNION{I} := \emptyset$; $\UNION{B} := \emptyset$; $\VAR{WaitList} :=
    \emptyset$;
    \cright{The first two are global lists.}

    \lnl{lnl:uca5-prune}
    \lIf{\ref{alg:prunecheck-uca5}($\mb{B}_0$, $\cl{C}_0$, $\varepsilon$, $\FC$, \VAR{WaitList})}
        \Return;
    \cright{On page~\pageref{alg:prunecheck-uca5}.}

    \While{$\VAR{WaitList} \neq \emptyset$}
    {
        \lnl{lnl:uca5-getnext}
        $(\mb{B}, \cl{C}) := \PRO{getNext}(\VAR{WaitList})$;
        \cright{Get the next element from the waiting list.}

        \lnl{lnl:uca5-split}
        $\cl{T} \equiv (\VAR{Splitter}, (\mb{B}_1, \dots, \mb{B}_k),
        \cl{C}, c)$ := $\ref{alg:branch-uca5}(\mb{B}, \cl{C})$;
        \cright{On page \pageref{alg:branch-uca5}.}

        \lIf{$\cl{T} = \emptyset$}{\CMD{continue while};}
        \cright{All points in $\mb{B}$ are solutions.}

        \For{$i := 1, \dots, k$}
        {
            \vspace{-1em}
            \cright{Do branching.}

            $\cl{C}_i := \cl{C}$;

            \If{$\VAR{Splitter} = \BS$ \CMD{and} $i > 1$}
            {
                \vspace{-1em}
                \cright{$\mb{B}_i$ does not contain the complementary box.}

                $\cl{C}_i := \cl{C}_i\setminus \set{c}$;
                \cright{The constraint $c$ is now redundant in $\mb{B}_i$
                (Theorem~\ref{thm:cbc-operator-empty}).}

                \If{$\cl{C}_i = \emptyset$}
                {
                    \vspace{-1em}
                    \cright{No running constraints.}

                    $\UNION{I} :=  \UNION{I} \cup \set{\mb{B}_i}$;
                    \cright{$\mb{B}_i$ is an inner box.}
                    \CMD{continue for};
                }
            }

            \lnl{lnl:uca5-prune2}
            $\ref{alg:prunecheck-uca5}(\mb{B}_i, \cl{C}_i, \varepsilon, \FC,
            \VAR{WaitList})$;
            \cright{On page~\pageref{alg:prunecheck-uca5}.}
        }
    }
\end{algorithm}

\begin{function}[!tb]
\caption{\ALG{PruneCheckUCA5}($\mb{B}$, $\cl{C}$, $\varepsilon$, $\FC$, \VAR{WaitList})}%
\label{alg:prunecheck-uca5}%
    \SetVline

    \lnl{lnl:uca5-dr}
    $\mb{B}' := \DR(\mb{B}, \cl{C})$;
    \cright{Prune the domains by using domain reduction operators.}

    \lIf{$\mb{B}' = \emptyset$} \Return{\CONST{true}};
    \cright{$\mb{B}$ is infeasible, the problem has been solved.}

    \lnl{lnl:uca5-check-eps}
    \If{there is no active variable in $\mb{B}'$ w.r.t. $\cl{C}$ and
    $\varepsilon$}
    {
        \vspace{-1em}
        \cright{Definition~\ref{def:active-var}.}

        \ref{alg:checkEpsilon}($\mb{B}'$, $\cl{C}$, $\FC$);
        \cright{On page~\pageref{alg:checkEpsilon}.}
        \Return{\CONST{true}};
    }

    \lnl{lnl:uca5-put}
    $\PRO{put}(\VAR{WaitList} \leftarrow (\mb{B}', \cl{C}))$;
    \cright{Put the current problem into the waiting list.}

    \Return{\CONST{false}}; \cright{The problem has not been solved yet.}
\end{function}

\begin{function}[!tb]
\caption{\ALG{SplitUCA5}($\mb{B}$, $\cl{C}$)}%
\label{alg:branch-uca5}%
    \SetVline
    \lnl{lnl:uca5-cb}
    \ForEach{$c \in \cl{C}$}
    {
        \vspace{-1em}
        \cright{Search for a constraint $c$ for which complementary boxing
        results in a reduction.}

        \lnl{lnl:uca5-cb-c}
        $\mb{CB}_c := \CB(\mb{B}, c)$;
        \cright{Enclose the negation of $c$ by using complementary boxing operators.}

        \lnl{lnl:uca5-c-empty}
        \If{$\mb{CB}_c = \emptyset$}
        {
            $\cl{C} := \cl{C} \setminus \{c\}$;
            \cright{$c$ is redundant in $\mb{B}$ (Theorem~\ref{thm:cbc-operator-empty}).}
            \CMD{continue for};
        }
        \lIf{$\mb{CB}_c \ne \mb{B}$}
        {
            \CMD{break for};
            \cright{Thus, $\mb{CB}_c \subset \mb{B}$.}
        }
    }

    \If{$\cl{C} = \emptyset$}
    {
        \vspace{-1em}
        \cright{No running constraints.}

        $\UNION{I} := \UNION{I} \cup \set{\mb{B}}$;
        \cright{$\mb{B}$ is an inner box.}
        \Return{$\emptyset$};
    }

    \lnl{lnl:uca5-bs}
    \If{$\mb{CB}_c \ne \mb{B}$}
    {
        $(\mb{B}_{1}$, $\dots$, $\mb{B}_{k}) := \BS(\mb{B}$, $\mb{CB}_c)$;
        $\VAR{Splitter} := \BS$;
        \cright{If $\BS$ did not fail, then $\mb{B}_{1} \supseteq \mb{CB}_c$.}
    }

    \lIf{$\BS$ \emph{failed}}
    {$\VAR{Splitter} := \DS$;}

    \lnl{lnl:uca5-ds}
    \lIf{$\VAR{Splitter} = \DS$}
    {$(\mb{B}_{1}, \dots, \mb{B}_{k}) := \DS(\mb{B})$;}
    \cright{Bisect $\mb{B}$, $k = 2$.}

    \lnl{lnl:uca5-split-ret}
    \Return{($\VAR{Splitter}$, $(\mb{B}_{1}, \dots, \mb{B}_{k})$, $\cl{C}$,
    $c$)};
\end{function}

\begin{remark}
In practice, equality constraints usually define surfaces, we then do not need
to perform the above Step~\ref{stp:uca5-2} for these constraints.
\end{remark}

The \ALG{UCA5} algorithm uses a waiting list, \VAR{WaitList}, to store the
subproblems waiting to be processed further. The elements can be retrieved
from, and be put to, \VAR{WaitList} by the functions $\PRO{getNext}$ and
$\PRO{put}$. \VAR{WaitList} can be handled as a queue or a stack. This allows
for the breadth-first search in the former case and the depth-first search in
the latter case.

In contrast to the \ALG{DMBC} and \ALG{DMBC$^+$} algorithms, the \ALG{UCA5}
algorithm restricts the $\DS$ operators at Line~\ref{lnl:uca5-ds} in Function
\ref{alg:branch-uca5} to dichotomizing a domain of a variable only if this
variable occurs in at least a running constraint. This avoids resulting in a
huge number of tiny boxes. The reason is that, in the \ALG{UCA5} algorithm,
constraints are removed from consideration whenever empty complementary boxes
are computed w.r.t. the constraints (see Line~\ref{lnl:uca5-c-empty} in
Function \ref{alg:branch-uca5}) and that, maybe, some variables no longer
appear in any running constraints. For simplicity, the interval (domain) with
the greatest width is selected for $\DS$ operators.

For efficiency, the $\BS$ operators at Line~\ref{lnl:uca5-bs} in Function
\ref{alg:branch-uca5} split along some facet of a complementary box only if
this produces sufficiently large boxes; the complementary box itself is
excepted. This estimation is done by using a predetermined parameter,
\PRO{fragmentation ratio}. After the splitting phase
(Line~\ref{lnl:uca5-split-ret} in Function \ref{alg:branch-uca5}), if the box
splitting operator was chosen and successful (i.e., $\VAR{Splitter} = \BS$),
then the first resulting box $\mb{B}_1$ contains the complementary box
$\mb{CB}_c$ and the constraint $c$ is always satisfied in all the other boxes
because of the complementariness of $\CB$ operators.

Function \ref{alg:prunecheck-uca5} (on page~\pageref{alg:prunecheck-uca5})
attempts to apply a $\DR$ operator to the input subproblem in order to reduce
the domains of the subproblem. If this returns an empty box, the subproblem has
no solutions. Afterwards, the procedure at Line~\ref{lnl:uca5-check-eps} in
Function \ref{alg:prunecheck-uca5} is to check if the input subproblem has no
active variable. If so, it uses a monotonic inclusion test, called $\FC$, to
check if the subproblem is either $\CONST{infeasible}$, $\CONST{feasible}$, or
$\CONST{unknown}$. If $\FC$ returns $\CONST{infeasible}$, the subproblem is
discarded. If $\FC$ returns $\CONST{unknown}$, the subproblem is classified as
\emph{undiscernible} w.r.t. $\varepsilon$ and $\FC$. In the other case, every
point in the domains of the subproblem is a solution. Although the monotonic
inclusion test $\FC$ in our implementation is a combination of $\DR$ and $\CB$
operators as described in Theorem~\ref{thm:fc-checker}, it is however not
restricted to this kind of monotonic inclusion test.

In Algorithm~\ref{alg:uca6}, we present a slightly generalized and improved
version of the \ALG{UCA6} algorithm -- a search technique proposed by
\citeN{SilaghiMC-SH-F:2001}. Basically, this version is the same as the
original version, but it is here improved by changing the order of the pruning
steps (Function \ref{alg:prunecheck-uca6}) such that each box is pruned before
being put into the waiting list ($\VAR{WaitList}$). This change reduces the
number of subproblems in the waiting list because some inconsistent subproblems
can be discarded sooner than that in the original version in
\cite{SilaghiMC-SH-F:2001}. This version is general enough to be used with the
heuristics in \cite{SilaghiMC-SH-F:2001}. Those heuristics are represented by a
generic function, called \PRO{getSplitType}, at
Line~\ref{lnl:uca6-getslittingtype} in Function \ref{alg:branch-uca6}.
Moreover, in this version, we make the stop condition more explicit at
Line~\ref{lnl:uca6-check-eps} in Function \ref{alg:prunecheck-uca6}. It is
important (for gaining in performance) to emphasize that checking if a domain
(i.e., an interval) is not wider than the predetermined precision $\varepsilon$
is only performed on the variables that occur in some running constraint,
because some constraints have become redundant. This detail has been omitted in
both \cite{SilaghiMC-SH-F:2001} and
\cite[Section~5.2.3]{SilaghiMC:Thesis:2002}.

\begin{algorithm}[!tb]
\caption{\ALG{UCA6} -- a new branch-and-prune search algorithm}%
\label{alg:uca6}%
    \SetVline
    \KwIn{a bounding box $\mb{B}_{0}$, a constraint set $\cl{C}_{0}$,
    $\varepsilon \in \Rp^n$, a monotonic inclusion test $\FC$.}

    \KwOut{an inner union approximation $\UNION{I}$, a boundary union
    approximation $\UNION{B}$.}

    $\UNION{I} := \emptyset$; $\UNION{B} := \emptyset$;
    $\VAR{WaitList} := \emptyset$;
    \cright{The first two are global lists.}

    \lnl{lnl:uca6-prune}
    \lIf{\ref{alg:prunecheck-uca6}($\mb{B}_0$, $\cl{C}_0$, $\set{\mb{B}_0,
    \dots, \mb{B}_0}$, $\varepsilon$, $\FC$, \VAR{WaitList})}
        \Return;
    \cright{Page~\pageref{alg:prunecheck-uca6}.}

    \While{$\VAR{WaitList} \neq \emptyset$}
    {
        \vspace{-1em}
        \cnext{A set  $\set{\mb{CB}_c \mid c \in \cl{C}}$ of memorized
        complementary boxes.}

        \lnl{lnl:uca6-getnext}
        $(\mb{B}, \cl{C}, \set{\mb{CB}_c \mid c \in \cl{C}}) :=
        \PRO{getNext}(\VAR{WaitList})$;

        \lnl{lnl:uca6-branch}
        $\cl{T} \equiv (\VAR{Spltr}, (\mb{B}_1, \dots, \mb{B}_k),
        \set{\mb{CB}_c \mid c \in \cl{C}}, b) :=
        \ref{alg:branch-uca6}(\mb{B}, \cl{C}, \set{\mb{CB}_c \mid c \in
        \cl{C}})$;

        \lIf{$\cl{T} = \emptyset$}{\CMD{continue while};}
        \cright{All points in $\mb{B}$ are solutions.}

        \For{$i := 1, \dots, k$}
        {
            \vspace{-1em}
            \cright{Do branching.}

            $\cl{C}_i := \cl{C}$;

            \If{$\VAR{Spltr} = \BS$ \CMD{and} $i > 1$}
            {
                $\cl{C}_i := \cl{C}_i\setminus \set{b}$;
                \cright{The constraint $b$ is now redundant in the box
                $\mb{B}_i$.}

                \If{$\cl{C}_i = \emptyset$}
                {
                    \vspace{-1em}
                    \cright{No running constraints.}

                    $\UNION{I} :=  \UNION{I} \cup \set{\mb{B}_i}$;
                    \cright{$\mb{B}_i$ is an inner box.}
                    \CMD{continue for};
                }
            }

            \lnl{lnl:uca6-prune2}
            $\ref{alg:prunecheck-uca6}(\mb{B}_i, \cl{C}_i,
            \set{\mb{CB}_c \mid c \in \cl{C}_{i}}, \varepsilon, \FC,
            \VAR{WaitList})$;
            \cright{On page~\pageref{alg:prunecheck-uca6}.}
        }
    }
\end{algorithm}

\begin{function}[!tb]
\caption{\ALG{PruneCheckUCA6}($\mb{B}$, $\cl{C}$, $\set{\mb{CB}_c \mid c \in
\cl{C}}$, $\varepsilon$, $\FC$, \VAR{WaitList})}%
\label{alg:prunecheck-uca6}%
    \SetVline

    \lnl{lnl:uca6-dr}
    $\mb{B}' := \DR(\mb{B}, \cl{C})$; \cright{Reduce domains.}

    \lIf{$\mb{B}' = \emptyset$} \Return{\CONST{true}};
    \cright{$\mb{B}$ is infeasible, the problem has been solved.}

    \lnl{lnl:uca6-check-eps}
    \If{there is no active variable in $\mb{B}'$ w.r.t. $\cl{C}$ and
    $\varepsilon$}
    {
        \vspace{-1em}
        \cright{Definition~\ref{def:active-var}.}
        \ref{alg:checkEpsilon}($\mb{B}'$, $\cl{C}$, $\FC$);
        \cright{On page~\pageref{alg:checkEpsilon}.}
        \Return{\CONST{true}};
    }

    \lnl{lnl:uca6-put}
    $\PRO{put}(\VAR{WaitList} \leftarrow (\mb{B}', \cl{C},
    \set{\mb{CB}_c \mid c \in \cl{C}}))$;
    \cright{Put the current problem into the waiting list.}

    \Return{\CONST{false}}; \cright{The problem has not been solved yet.}
\end{function}

\begin{function}[!tb]
\caption{\ALG{SplitUCA6}($\mb{B}$, $\cl{C}$, \set{\mb{CB}_c \mid c \in \cl{C}})}%
\label{alg:branch-uca6}%
    \SetVline
    \lnl{lnl:uca6-cb}
    \ForEach{$c \in \cl{C}$}
    {
        \lnl{lnl:uca6-cb-c}
        $\mb{CB}_c := \CB(\mb{B} \cap \mb{CB}_c, c)$;

        \lnl{lnl:uca6-c-empty}
        \lIf{$\mb{CB}_c = \emptyset$}{$\cl{C} := \cl{C} \setminus \{c\}$;}
        \cright{$c$ is redundant in $\mb{B}$ (Theorem~\ref{thm:cbc-operator-empty}).}
    }

    \If{$\cl{C} = \emptyset$}
    {
        \vspace{-1em}
        \cright{No running constraints.}

        $\UNION{I} := \UNION{I} \cup \set{\mb{B}}$;
        \cright{$\mb{B}$ is an inner box.}
        \Return{$\emptyset$};
    }

    \lnl{lnl:uca6-getslittingtype}
    $\VAR{Splitter} := \PRO{getSplitType}()$;
    \cright{Get a splitting mode, heuristics can be used.}

    \If{$\VAR{Splitter} = \BS$}
    {
        \vspace{-1em}
        \cright{The splitting mode is box splitting.}

        $\mb{CB}_b := \PRO{chooseTheBest}(\{\mb{CB}_c \mid c \in \cl{C}\})$\;

        $(\mb{B}_{1}$, $\dots$, $\mb{B}_{k}) := \BS(\mb{B}$, $\mb{CB}_b)$;
        \cright{If box splitting did not fail, then $\mb{B}_{1} \supseteq
        \mb{CB}_b$.}

        \lIf{$\BS$ \emph{failed}}
        {$\VAR{Splitter} := \DS$;}
    }

    \lnl{lnl:uca6-ds}
    \lIf{$\VAR{Splitter} = \DS$}
    {$(\mb{B}_{1}, \dots, \mb{B}_{k}) := \DS(\mb{B})$;}
    \cright{Bisect $\mb{B}$, $k = 2$.}

    \lnl{lnl:uca6-split-ret}
    \Return{($\VAR{Splitter}$, $(\mb{B}_{1}, \dots, \mb{B}_{k})$,
    $\set{\mb{CB}_c \mid c \in \cl{C}}$, $b$)};
\end{function}

In fact, the \ALG{UCA5} algorithm is a simplification of the \ALG{UCA6}
algorithm. Thus, these two algorithms are very much similar and only different
at the followings:

\begin{itemize}
    \item At Line~\ref{lnl:uca6-prune} and Line~\ref{lnl:uca6-prune2}: Function
    \ref{alg:prunecheck-uca6} takes not only the same arguments as Function
    \ref{alg:prunecheck-uca5} does but also a list of complementary boxes
    computed at the parent of the current search node. Each box in this list is
    a complementary box w.r.t. a running constraint. At the beginning
    (Line~\ref{lnl:uca6-prune}), this list consists of boxes that equal to the
    initial bounding box.

    \item At Line~\ref{lnl:uca6-getnext} and Line~\ref{lnl:uca6-put}: Each
    element in the waiting list $\VAR{WaitList}$ of \ALG{UCA6} contains not
    only a domain box and a set of running constraints but also a list of
    complementary boxes computed at the parent of the current search node.

    \item At Line~\ref{lnl:uca6-branch}: Function \ref{alg:branch-uca6} takes
    not only a box playing the role of domains and a set of running constraints
    but also a list of complementary boxes gotten from the waiting list. This
    function returns not only the used splitting type and the list of boxes as
    Function \ref{alg:branch-uca5} does but also a \emph{new} list of
    complementary boxes (because some constraints may have been removed).

    \item From Line~\ref{lnl:uca6-cb} to Line~\ref{lnl:uca6-c-empty}: This is
    the main difference between the \ALG{UCA6} algorithm and the \ALG{UCA5}
    algorithm. While the \ALG{UCA5} algorithm only finds a complementary box
    that is strictly contained in the bounding box, the \ALG{UCA6} algorithm
    computes complementary boxes for every constraints and then chooses the
    best.

    \item From Line~\ref{lnl:uca6-getslittingtype} to
    Line~\ref{lnl:uca5-split-ret}: Since the \ALG{UCA6} algorithm has just
    computed complementary boxes for every constraints, it chooses the smallest
    complementary box $\mb{CB}_b$ based on the volume which was computed for
    the constraint $b$. The constraint $b$ will be used for box splitting
    operators. In the \ALG{UCA5} algorithm, the first-found complementary box
    which is strictly contained in the bounding box will be used for box
    splitting operators. Notice that the \ALG{UCA5} algorithm does not need an
    additional amount of memory to remember the computed complementary boxes in
    the waiting list $\VAR{WaitList}$.
\end{itemize}

Of course, both the \ALG{UCA5} and \ALG{UCA6} algorithms are instances of the
$\ALG{BnPSearch}$ algorithm. They compute inner and boundary union
approximations, $\UNION{I}$ and $\UNION{B}$ respectively. These two union
approximations are disjoint. Thus, we can obtain an outer union approximation
by setting $\UNION{O} := \UNION{I} \cup \UNION{B}$.

\begin{theorem}
\label{thm:uca5n6}%
Given a monotonic inclusion test $\FC$ and a positive precision (vector)
$\varepsilon$. Both the \ALG{UCA5} and \ALG{UCA6} algorithms terminate and
provide inner and boundary union approximations, $\UNION{I}$ and $\UNION{B}$
respectively, at the precision $\varepsilon$ with respect to the monotonic
inclusion test $\FC$ (see Definition~\ref{def:interval-precision}).
\end{theorem}

\begin{proof}
No solution is lost because of the correctness of $\DR$ operators
(Definition~\ref{def:dr-operator}). Moreover, all points of any boxes in
$\UNION{I}$ are sound solutions. That is due to the complementariness of $\CB$
operators (Definition~\ref{def:cbc-operator}) and
Theorem~\ref{thm:cbc-operator-empty}. Therefore, $\UNION{I}$ and $\UNION{B}$
are inner and boundary union approximations of the solution set, respectively.

If $\mb{B}$ is a box in $\UNION{B}$, then
\begin{itemize}
    \item The box $\mb{B}$ has no active variable w.r.t. $\varepsilon$ and the
    running constraints $\cl{C}$ (at Line~\ref{lnl:uca6-check-eps} of Function
    \ref{alg:prunecheck-uca5} and Function \ref{alg:prunecheck-uca6});

    \item Function \ref{alg:checkEpsilon} (on page~\pageref{alg:checkEpsilon})
    returns $\CONST{unknown}$ (i.e., $\FC(\mb{B}, \cl{C}) = \CONST{unknown}$).
\end{itemize}
If a variable $x_i$, of which the domain is $\proj{\mb{B}}{x_i}$, is in
$\vars(\cl{C})$ (the set of variables of $\cl{C}$), then the width
$\Wid(\proj{\mb{B}}{x_i})$ is not greater than $\varepsilon_i$. If a variable
$x_j$ is not in $\vars(\cl{C})$, we then split the interval
$\proj{\mb{B}}{x_j}$ of $\mb{B}$ into intervals whose widths are not greater
than $\varepsilon_j$. Eventually, we obtain a number of $\varepsilon$-bounded
boxes whose union is $\mb{B}$. It follows from the properties of a monotonic
inclusion test (Definition~\ref{def:fc-checker}) that each obtained
$\varepsilon$-bounded box $\mb{B}'$ satisfies the property $\FC(\mb{B}',
\cl{C}) = \FC(\mb{B}, \cl{C}) = \CONST{unknown}$, since the projections
$\proj{\mb{B}'}{\vars(\cl{C})} = \proj{\mb{B}}{\vars(\cl{C})}$.
\end{proof}

Roughly speaking, the larger the union $\UNION{I}$ is and the smaller the union
$\UNION{B}$ is, the more accurate the solution algorithm is.

\section{Compacting the Representation of Solutions}
\label{sec:search-impr-algorithm}

\subsection{Controlling the Reduction of Small Domains}
\label{sec:stop-reduction}%

When using the \ALG{UCA5} algorithm or the \ALG{UCA6} algorithm to solve NCSPs
with continuums of solutions, we observe that a better alignment of boxes near
the boundary of the solution set can be obtained by finely controlling the
application of domain reduction and complementary boxing operators. In
particular, if a variable $x_i$ of a subproblem is \emph{inactive} (see
Definition~\ref{def:active-var}), then the domain reduction or complementary
boxing should not be enforced on the variable $x_i$. This is to obtain better
alignments of contiguous boxes and the computational performance.

A domain reduction operator (respectively, a complementary boxing operator)
that only reduce the domains of active variables is called a
\emph{restricted-dimensional domain reduction operator} (respectively, a
\emph{restricted-dimensional complementary boxing operator}). We denote by
$\DRrd$ (respectively, $\CBrd$) the restricted-dimensional domain reduction
operator (respectively, the restricted-dimensional complementary boxing
operator) obtained from the normal domain reduction operator $\DR$
(respectively, the normal complementary boxing operator $\CB$) by enforcing the
reduction only on active variables.

When local consistency techniques are enforced in order to obtain the effect of
domain reduction (i.e., they play the role of domain reduction operators),
restricted-dimensional operators amount to enforcing the local consistency
techniques only for active variables. The recent algorithms for achieving box
consistency, hull consistency and $k\BB$-consistency can be easily modified to
adopt the above idea about restricted-dimensional operators, for example, by
ignoring any procedure involving the inactive variables. In case a domain
reduction or complementary boxing operator cannot be modified to adopt the idea
of reducing only on active variables, we can apply it normally and then restore
the domains of inactive variables to the initial domains. In this case, the
gain is only in the (better) alignment of contiguous boxes, but not in the
performance.

Fortunately, the implementation of box consistency in a well-known product
called \PRO{ILOG Solver} \cite{ILOGSolver-6.0:2003} supports the above idea
about reducing only on active variables. It can be done by simply passing only
active variables (\texttt{X}) to the function \texttt{IloGenerateBounds} when
we need to generate narrower bounds on \texttt{X}.

\begin{figure}[!tb]
    \centerline{\includegraphics[width=11.5cm]{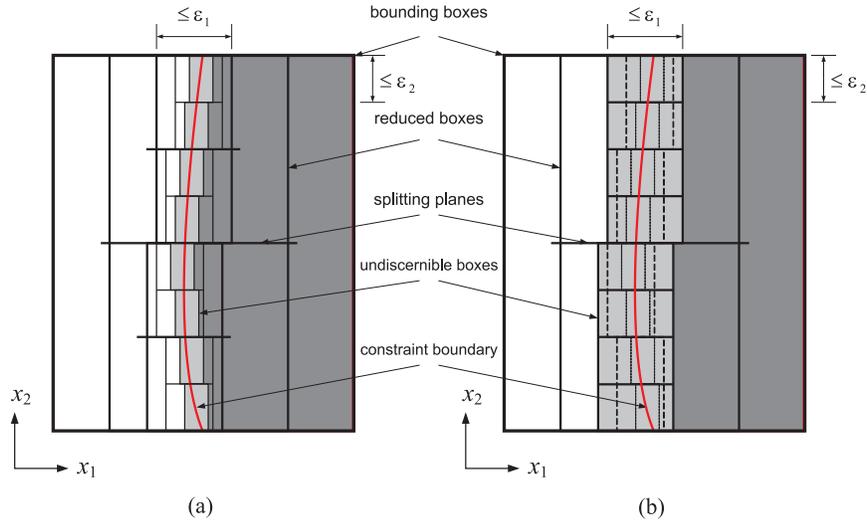}}
\caption[Normal domain reductions and restricted-dimensional domain
reductions]{An example of normal domain reductions and restricted-dimensional
domain reductions at different levels: (a) all variables ($x_1$ and $x_2$) are
considered for the over-reduction; (b) only the active variable ($x_2$) is
considered for the reduction. The dark grey regions are inner boxes, the light
grey regions are undiscernible boxes.}
\label{fig:stop-reduction}%
\end{figure}

An illustration of the difference between the effect of a normal domain
reduction ($\DR$) operator and the effect of a restricted-dimensional domain
reduction ($\DRrd$) operator in the solving process is presented in
Figure~\ref{fig:stop-reduction}. In this example, although the normal $\DR$
operator produces more accurate output than the $\DRrd$ operator does, it has
to spend much time on making the boundary region narrower than the allowed
tolerance $\varepsilon_1$. In practice, this is unnecessary since real world
applications mainly focus on the inner boxes, the boxes near the boundary are
often unsafe for the further exploration in the applications, as shown in our
experiments (see Section~\ref{sec:search-experiment}). Moreover, the number of
boxes (15 inner boxes and 8 undiscernible boxes) resulting from the application
of the $\DR$ operator is often higher than the number of boxes (3 inner boxes
and 8 undiscernible boxes) resulting from the application of the $\DRrd$
operator. Moreover, the contiguous boxes obtained by using the $\DRrd$ operator
are often aligned; hence, a geometrically compacting technique can work on them
efficiently to get a concise representation of the solution set, as shown
below.

\subsection{Compact Representation of Solutions}
\label{sec:compact-rept}%

Once the effect of better alignments is obtained, the question is how such a
set of aligned boxes can be compacted into a smaller set. We propose to use the
\emph{extreme vertex representation} (EVR) of orthogonal polyhedra for this
purpose. The extreme vertex representation was first proposed by
\citeN{AguileraA-A:1997} for the three-dimensional space, and was later
extended to the $n$-dimensional space in
\cite{BournezO-M-P:1999,BournezO-M:2000}. The basic idea is that the union of
disjoint boxes delivered by a box-covering solver defines an \emph{orthogonal
polyhedron} for which an improved representation can be generated. An
orthogonal polyhedron can be naturally represented as the union of disjoint
boxes (by enumerating the boxes and their vertices). That representation is
called the \emph{disjoint box representation} (DBR) in computational geometry.
The EVR is a way of compacting the DBR (see \cite{AguileraA:Thesis:1998} and
\cite{BournezO-M-P:1999,BournezO-M:2000}). Roughly speaking, in the EVR, the
extreme vertices of an orthogonal polyhedron are identified and stored in a
compact manner such that no information is lost w.r.t. the representation of
the polyhedra. Moreover, when converting from EVR back to DBR, the obtained DBR
are often more compact than the initial DBR.

\begin{figure}[!tb]
    \centerline{\includegraphics[width=10cm]{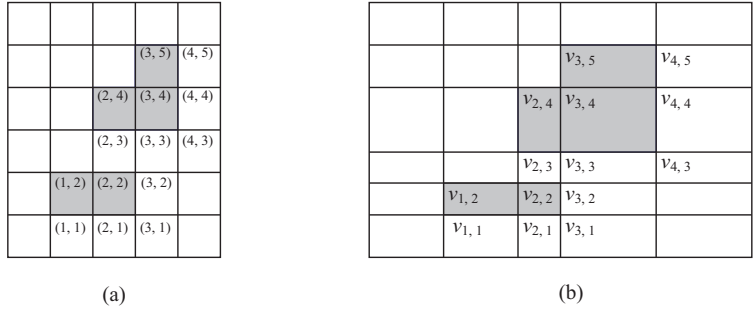}}
\caption[Examples of a griddy polyhedron and an orthogonal polyhedron]{Examples
of a griddy polyhedron and an orthogonal polyhedron: (a) a griddy polyhedron
made of the vertex indices of (b) an orthogonal polyhedron.}
\label{fig:orthogonal-polyhedra}%
\end{figure}

We recall here basic concepts in the theory of extreme vertex representation.
The reader can find more details in \cite{BournezO-M:2000}. These concepts are
sufficient to be presented for \emph{griddy polyhedra}\footnote{A griddy
polyhedron is the union of some unit hypercubes with integer-valued vertices
(see \cite{BournezO-M-P:1999}).} because the results on general orthogonal
polyhedra can be easily obtained from the results on griddy polyhedra by
mapping the multidimensional array of vertex indices of the orthogonal
polyhedra to the multidimensional array of vertices of griddy polyhedra (see
Figure~\ref{fig:orthogonal-polyhedra}). In fact, they do not depend on the
orthogonality of the underlying basis.

For simplicity, polyhedra are assumed to live in $\mb{X} = \rintv{0}{m}^d
\subseteq \R^d$ (the results also hold for $\mb{X} = \Rp^d)$. Let $\cl{G} =
\seq{0, 1, \dots, m-1}^d \subseteq \N^d$ be a grid of integer points. For every
point $x \in \mb{X}$, \ipart{x} denotes the grid point corresponding to the
(componentwise) integer part of $x$. The unit box associated with a grid point
$x = (x_1, \dots, x_d)^\TT \in \cl{G}$ is the box $\mb{B}(x) = \rintv{x_1}{x_1
+ 1} \times \dots \times \rintv{x_d}{x_d + 1}$. The set of all unit boxes is
denoted by $\cl{B}$. A griddy polyhedron $P$ is the set closure of the union of
some unit boxes, or can be viewed as a set of grid points.

\begin{definition}[Color Function]
\label{def:color-function} Let $P$ be a griddy polyhedron. The \emph{color
function} $\PRO{color}: \mb{X} \to \{0,1\}$ is defined as follows: if $x$ is a
grid point then $\PRO{color}(x) = 1 \IFF \mb{B}(x) \subseteq P$; otherwise,
$\PRO{color}(x) = \PRO{color}(\ipart{x})$.
\end{definition}

\begin{figure}[!htb]
    \centerline{\includegraphics[width=7cm]{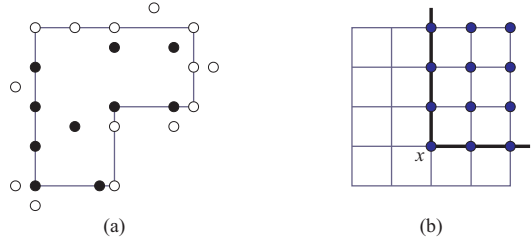}}
\vspace{-0.5em}%
\caption[A griddy polyhedron with samples of colors]{A griddy polyhedron: (a)
sample colors defined by the color function; (b) the forward cone
$x^{\triangleleft} \equiv \set{y \in \cl{G} \mid x \le y}$.}
\label{fig:evr-color}%
\end{figure}

We say that a grid point $x$ is black (respectively, white) and that
$\mb{B}(x)$ is full (respectively, empty) when $\PRO{color}(x) = 1$
(respectively, $\PRO{color}(x) = 0$). Figure~\ref{fig:evr-color}a illustrates
the color function for griddy polyhedra. Figure~\ref{fig:evr-color}b
illustrates the concept of a \emph{forward cone} based at $x \in \cl{G}$, which
is defined as $x^{\triangleleft} \equiv \set{y \in \cl{G} \mid x \le y}$.

A \emph{canonical representation} scheme for $2^{\cl{B}}$ (or $2^{\cl{G}}$) is
a set $\cl{E}$ of syntactic objects such that there is some bijective function
$\Psi : \cl{E}\to 2^{\cl{B}}$; that is, every polyhedron has a unique
representation. The most simple representation scheme is to explicitly
enumerate the values of the color function on every grid point; hence, it needs
a $d$-dimensional array of bits with $m^d$ entries. Another simple
representation is the vertex representation that consists of the set $\set{(x,
\PRO{color}(x)) \mid x \tn{ is a vertex}}$. This is however still verbose.
Hence, it is desired to store only important vertices only. The following
definition identifies those important vertices.

\begin{definition}[Extreme Vertex]
\label{def:extreme-vetex}%
A grid point $x$ is called an \emph{extreme vertex} of a griddy polyhedron $P$
if the number of black grid points in $\cl{N}(x) = \set{x_1 - 1, x_1} \times
\dots \times \set{x_d - 1, x_d}$ is odd ($\cl{N}(x)$ is called the
\emph{neighborhood} of $x$).
\end{definition}

Let $\oplus$ denote the exclusive-or (XOR) operation: $p \oplus q = (p \wedge
\lnot q) \vee (p \wedge \lnot q)$. The $\oplus$ operation on sets is defined by
$A \oplus B = \set{x \mid (x \in A) \oplus (x \in B)}$. The set of extreme
vertices (together with the $\oplus$ operation) makes a canonical
representation of griddy polyhedra as follows \cite[Theorem~1 \&
2]{BournezO-M:2000}.

\begin{theorem}[Extreme Vertex Representation]
\label{thm:evr-representation}%
For any griddy polyhedron $P$, there is a unique set $V$ of grid points in
$\cl{G}$ such that $P = \bigoplus_{x \in V} x^{\triangleleft}$. Moreover, the
set $V$ is the set of all extreme vertices of $P$ and this is a canonical
representation.
\end{theorem}

\begin{figure}[!tb]
    \centerline{\includegraphics[width=10.5cm]{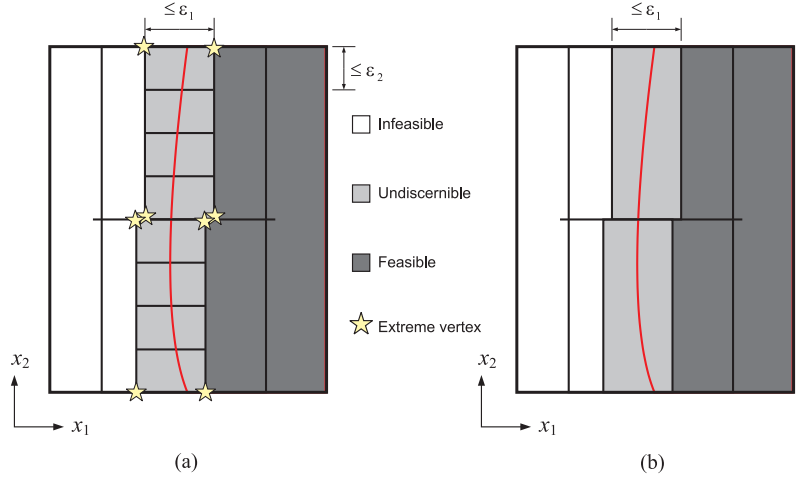}}
\caption[An example of extreme vertices of union approximations]{The use of
extreme vertex representations compacts the boundary union approximation in
Figure~\ref{fig:stop-reduction}b.}
\label{fig:evr}%
\end{figure}

Figure~\ref{fig:evr} illustrates how the application of the concept of extreme
vertex representations compacts union approximations. The eight light grey
boxes of the boundary union approximation in Figure~\ref{fig:evr}a can be
concisely and equivalently represented by the two light grey boxes in
Figure~\ref{fig:evr}b. Theorem~\ref{thm:evr-representation} shows that any
griddy polyhedron can be canonically represented by the set of its extreme
vertices (and their colors).

An effective transformation between DBR and EVR was proposed for low
dimensional or small-size (i.e., $m$ is small) polyhedra
\cite{AguileraA:Thesis:1998,BournezO-M:2000}. For example, in the
three-dimensional space, the average experimental time complexity of converting
an EVR to a DBR is far less than quadratic but slightly greater than linear in
the number of extreme vertices \cite{AguileraA:Thesis:1998}. Results in
\cite{BournezO-M:2000} also imply that, in a fixed dimension, the time
complexity of converting a DBR to an EVR by using the XOR operator is linear in
the number of boxes in DBR. We then propose to exploit the ideas of these
effective transformation schemes to produce a compact representation of
contiguous aligned boxes. This process works as follows:

\begin{enumerate}
    \item Produce a better alignment of the boxes along the boundaries of
    constraints. This is done by preventing the unnecessary application of
    reduction operators over inactive variables.
    Figure~\ref{fig:stop-reduction} shows the effect of better alignment
    obtained for a set of nearly aligned boxes of an undiscernible
    approximation. The original set of eight small boxes
    (Figure~\ref{fig:stop-reduction}a) reduces to two groups of four aligned
    boxes (Figure~\ref{fig:stop-reduction}b) without altering the predetermined
    precision.

    \item The \ALG{Combination} function: The set of aligned boxes in each
    group, $\cl{S}_{1}$, is converted to EVR and then back to DBR to get a set
    of combined boxes, $\cl{S}_{2}$ (containing only one box in this case). Due
    to the properties of EVR, $\cl{S}_{2}$ is often more concise than
    $\cl{S}_{1}$. Figure~\ref{fig:evr} shows how this conversion procedure
    reduces eight boxes to two boxes.
\end{enumerate}

The above conversion procedure can theoretically be applied in any dimension.
Due to the efficiency of EVR in low dimension, we however restrict its
application to very low dimensional small-size regions of the search space in
our implementation (see Section \ref{sec:new-adv-search-algorithm}). The
running time for this conversion is hence near zero.

\subsection{An Improved Branch-and-Prune Search Algorithm}
\label{sec:new-adv-search-algorithm}

We present in Algorithm~\ref{alg:uca6plus} an improved version of the
\ALG{UCA5} and \ALG{UCA6} algorithms, which is called \ALG{UCA6$^+$}. It also
takes as input an NCSP, $\cl{P} \equiv (\cl{V}, \cl{C}_0, \mb{B}_0)$, and
returns an inner union approximation $\UNION{I}$ and a boundary union
approximation $\UNION{B}$ of the solution set of $\cl{P}$. Roughly speaking,
the \ALG{UCA6$^+$} algorithm uses restricted-dimensional versions of domain
reduction and complementary boxing operators instead of normal versions in
order to produce the effect of better alignment (and also to gain in
performance), and then uses the conversion between the extreme vertex
representation and disjoint box representation to get a compact representation
of union approximations. The main processes of the \ALG{UCA6$^+$} algorithm are
similar to the three main processes of the the \ALG{UCA5} and \ALG{UCA6}
algorithm, but the \ALG{UCA6$^+$} algorithm uses restricted-dimensional domain
reduction ($\DRrd$) operators in place of domain reduction ($\DR$) operators
(see Line~\ref{lnl:uca6p-dr} in Function \ref{alg:prunecheck-uca6p}). The
\ALG{UCA6$^+$} algorithm also uses restricted-dimensional complementary boxing
($\CBrd$) operators in place of complementary boxing ($\CB$) operators (see
Line~\ref{lnl:uca6p-cb} in Function \ref{alg:split-uca6p}).


\begin{algorithm}[!b]
\caption{\ALG{UCA6$^+$} -- a new branch-and-prune search algorithm}%
\label{alg:uca6plus}%
    \SetVline
    \KwIn{a box $\mb{B}_{0}$, a constraint set $\cl{C}_{0}$,
    $\varepsilon \in \Rp^n$, a monotonic inclusion test $\FC$, $D_{\tn{stop}}$.}

    \KwOut{an inner union approximation $\UNION{I}$, a boundary union
    approximation $\UNION{B}$.}

    $\UNION{I} := \emptyset$; $\UNION{B} := \emptyset$;
    $\VAR{WaitList} := \emptyset$;
    \cright{The first two are global lists.}

    \lnl{lnl:uca6p-prune}
    \lIf{\ref{alg:prunecheck-uca6p}($\mb{B}_0$, $\cl{C}'_0$, $\emptyset$,
    $\varepsilon$, $\FC$, $\VAR{WaitList}$, $D_{\tn{stop}}$)}
    {
        \Return;
    }
    \cright{On page~\pageref{alg:prunecheck-uca6p}.}

    \While{$\VAR{WaitList} \neq \emptyset$}
    {
        \cnext{A set  $\set{\mb{CB}_c \mid c \in \cl{C}}$ of complementary
        boxes that were \emph{optionally} memorized.}
        \lnl{lnl:uca6p-getnext}
        $(\mb{B}, \cl{C}, \set{\mb{CB}_c \mid c \in \cl{C}'}) :=
        \PRO{getNext}(\VAR{WaitList})$;
        \cright{$\cl{C}' \subseteq \cl{C}$, $\mb{CB}_c \subset \mb{B}$.}

        \lnl{lnl:uca6p-branch}
        $(\VAR{Spltr}, (\mb{B}_{1}, \dots, \mb{B}_{k}), \set{\mb{CB}_c
        \mid c \in \cl{C}'}, \cl{C}, b) := \ref{alg:split-uca6p}(\mb{B}, \cl{C},
        \set{\mb{CB}_c \mid c \in \cl{C}'})$;

        \lIf{\ref{alg:split-uca6p} returns $\emptyset$}
            \CMD{continue while};

        \For{$i := 1, \dots, k$}
        {
            \vspace{-1em}
            \cright{Do branching.}

            $\cl{C}_i := \cl{C}$; $\cl{C}'_{i} := \cl{C}'$\;

            \If{$\VAR{Spltr} = \BS$ \CMD{and} $i > 1$}
            {
                $\cl{C}_i := \cl{C}_i\setminus \{ b \}$;
                $\cl{C}'_{i} := \cl{C}'_{i} \setminus \set{b}$;
                \cright{$b$ is now redundant in the box $\mb{B}_{i}$
                (Theorem~\ref{thm:cbc-operator-empty}).}

                \If{$\cl{C}_i = \emptyset$}
                {
                    \vspace{-1em}
                    \cright{No running constraints.}

                    $\UNION{I} :=  \UNION{I} \cup \set{\mb{B}_i}$;
                    \cright{$\mb{B}_i$ is an inner box.}
                    \CMD{continue for}\;
                }
            }

            \lnl{lnl:uca6p-remove-cb}
            \lForEach{$c \in \cl{C}'_{i}$}
            {
                \lIf{$\mb{B}_i \subseteq \mb{CB}_c$}
                {
                    $\cl{C}'_{i} := \cl{C}'_{i} \setminus \set{c}$\;
                }
            }

            \lnl{lnl:uca6p-prune2}
            \ref{alg:prunecheck-uca6p}($\mb{B}_i$, $\cl{C}_i$,
            $\set{\mb{B}_i \cap \mb{CB}_c \mid c \in \cl{C}'_{i}}$,
            $\varepsilon$, $\FC$, $\VAR{WaitList}$, $D_{\tn{stop}}$);
        }
    }
\end{algorithm}

\begin{function}[!tb]
\caption{\ALG{PruneCheckUCA6$^+$}($\mb{B}$, $\cl{C}$, $\set{\mb{CB}_c \mid c
\in \cl{C}'}$, $\varepsilon$, $\FC$, \VAR{WaitList}, $D_{\tn{stop}}$)}%
\label{alg:prunecheck-uca6p}%
    \SetVline

    \lnl{lnl:uca6p-dr}
    $\mb{B}' := \DRrd(\mb{B}, \cl{C})$;
    \cright{Restricted-dimensional domain reduction.}

    \lIf{$\mb{B}' = \emptyset$} \Return{\CONST{true}};
    \cright{$\mb{B}$ is infeasible, the problem has been solved.}

    \lnl{lnl:uca6p-check-eps}
    \If{there is no active variable in $\mb{B}'$ w.r.t. $\cl{C}$ and
    $\varepsilon$}
    {
        \vspace{-1em}
        \cright{Definition~\ref{def:active-var}.}

        \ref{alg:checkEpsilon}($\mb{B}'$, $\cl{C}$, $\FC$);
        \cright{On page~\pageref{alg:checkEpsilon}.}
        \Return{\CONST{true}};
    }

    \lnl{lnl:uca6p-resort}
    \If{there are at most $D_{\tn{stop}}$ active variables in $\mb{B}'$}
    {
        \vspace{-1em}
        \cnext{Resort to another technique.}

        \lnl{lnl:uca6p-dimstopsolver}
        ($\UNION{I}(\mb{B}', \cl{C})$, $\UNION{B}(\mb{B}', \cl{C})$) :=
        $\ALG{DimStopSolver}(\mb{B}', \cl{C}, \varepsilon, \FC, \DRrd, \CBrd)$;

        \cnext{$\ALG{Combination}(.)$ does the conversions DBR $\to$ EVR
        $\to$ DBR in a $D_{\tn{stop}}$-dimensional space.}

        $\UNION{I} := \UNION{I} \cup \ALG{Combination}(\UNION{I}(\mb{B}', \cl{C}))$;
        \cright{Store in the global list of feasible boxes.}

        $\UNION{B} := \UNION{B} \cup \ALG{Combination}(\UNION{B}(\mb{B}', \cl{C}))$;
        \cright{Store in the global list of undiscernible boxes.}

        \Return{\CONST{true}}; \cright{The problem has been solved.}
    }

    \lnl{lnl:uca6p-remove-cb2}
    \lForEach{$c \in \cl{C}'$}
    {
        \lIf{$\mb{B}' \subseteq \mb{CB}_c$}
        {
            $\cl{C}' := \cl{C}' \setminus \set{c}$\;
        }
    }

    \lnl{lnl:uca6p-put}
    $\PRO{put}(\VAR{WaitList} \leftarrow (\mb{B}', \cl{C}, \set{\mb{B}' \cap
    \mb{CB}_c \mid c \in \cl{C}'}))$;
    \cright{Put the problem into the waiting list.}

    \Return{\CONST{false}}; \cright{The problem has not been solved yet.}
\end{function}

\begin{function}[!tb]
\caption{\ALG{SplitUCA6$^+$}($\mb{B}$, $\cl{C}$, \set{\mb{CB}_c \mid c \in \cl{C}'})}%
\label{alg:split-uca6p}%
    \SetVline
    \lnl{lnl:uca6p-choice-c}
    Choose an arbitrary subset $\cl{C}'' \subseteq \cl{C}$;
    \cright{$\cl{C}''$ is a set of constraints to be used with the
    $\CBrd$ operator.}

    \lnl{lnl:uca6p-cb}
    \ForEach{$c \in \cl{C}' \cup \cl{C}''$}
    {
        \uIf{$c \in \cl{C}' \cap \cl{C}''$}
        {
            \lnl{lnl:uca6p-cb-c1}
            $\mb{CB}_c := \CBrd(\mb{B} \cap \mb{CB}_c, c)$\;
        }
        \uElseIf{$c \in \cl{C}''$}
        {
            \vspace{-1em}
            \cright{$c \notin \cl{C}'$.}

            \lnl{lnl:uca6p-cb-c2}
            $\mb{CB}_c := \CBrd(\mb{B}, c)$;
            $\cl{C}' := \cl{C}' \cup \set{c}$\;
        }
        \Else
        {
            \vspace{-1em}
            \cright{$c \in \cl{C}'$, $c \notin \cl{C}''$.}

            \lnl{lnl:uca6p-cb-c3}
            $\mb{CB}_c := \mb{B}\cap \mb{CB}_c$\;
        }
        \lIf{$\mb{CB}_c = \emptyset$}{$\cl{C} := \cl{C} \setminus \set{c}$;}
        \cright{$c$ is now redundant in $\mb{B}$
        (Theorem~\ref{thm:cbc-operator-empty}).}

        \lIf{$\mb{CB}_c = \emptyset$ \CMD{or} $\mb{CB}_c = \mb{B}$}
        {
            $\cl{C}' := \cl{C}' \setminus \set{c}$\;
        }
    }

    \lnl{lnl:uca6p-c-empty}
    \If{$\cl{C} = \emptyset$}
    {
        \vspace{-1em}
        \cright{No running constraints.}

        $\UNION{I} := \UNION{I} \cup \set{\mb{B}}$;
        \cright{$\mb{B}$ is an inner box.}
        \Return{$\emptyset$}\;
    }

    \lnl{lnl:uca6p-getsplittype}
    $\VAR{Splitter} := \PRO{getSplitType}()$;
    \cright{Get a splitting mode, heuristics can be used.}

    \lnl{lnl:uca6p-bs}
    \If{$\VAR{Splitter} = \BS$}
    {
        \vspace{-1em}
        \cright{The splitting mode is box splitting.}

        \lnl{lnl:uca6p-choosethebest}
        $\mb{CB}_b := \PRO{chooseTheBest}(\set{\mb{CB}_c \mid c \in \cl{C}'})$\;

        $(\mb{B}_{1}$, $\dots$, $\mb{B}_{k}) := \BS(\mb{B}$, $\mb{CB}_b)$;
        \cright{If box splitting did not fail, then $\mb{B}_{1} \supseteq \mb{CB}_b$.}

        \lIf{$\cl{C}' = \emptyset$  \textbf{or} $\BS$ \emph{failed}}
        {$\VAR{Splitter} := \DS$;}
    }

    \lnl{lnl:uca6p-ds}
    \lIf{$\VAR{Splitter} = \DS$}
    {$(\mb{B}_{1}, \dots, \mb{B}_{k}) := \DS(\mb{B})$;}
    \cright{Bisect $\mb{B}$, $k = 2$.}

    \lnl{lnl:uca6p-split-ret}
    \Return{($\VAR{Splitter}$, $(\mb{B}_{1}, \dots, \mb{B}_{k})$,
    $\set{\mb{CB}_c \mid c \in \cl{C}'}$, $\cl{C}$, $b$)};
\end{function}


The \ALG{UCA6$^+$} algorithm does not compute complementary boxes for all
running constraints as the \ALG{UCA6} algorithm does. Instead, it allows users
to predefine a policy to choose a subset $\cl{C}''$ of $\cl{C}$, of which the
constraints are enforced with $\CBrd$ operators (see
Line~\ref{lnl:uca6p-choice-c} in Function \ref{alg:split-uca6p}). A simple
policy is to choose either all the constraints of $\cl{C}$ or a fixed number of
constraints in $\cl{C}$. A more complicated and dynamic policy based on the
pruning efficiency can be used. The set of constraints to be considered in the
computation of complementary boxes -- which is done by using $\CBrd$ operators
(at Line~\ref{lnl:uca6p-cb-c1} and Line~\ref{lnl:uca6p-cb-c2}) or by
intersecting with the memorized complementary boxes (at
Line~\ref{lnl:uca6p-cb-c3}) -- is thus the union $\cl{C}' \cup \cl{C}''$, where
$\cl{C}'$ is the set of constraints associated with the memorized complementary
boxes (see Line~\ref{lnl:uca6p-getnext}, \ref{lnl:uca6p-branch},
\ref{lnl:uca6p-prune2}, \ref{lnl:uca6p-put} and \ref{lnl:uca6p-split-ret} in
Algorithm~\ref{alg:uca6plus}). Notice that the set $\cl{C}'$ is only a subset
of $\cl{C}$, in general.

The \ALG{UCA6$^+$} algorithm uses the same functions \PRO{getSplitType} and
\PRO{chooseTheBest} as the \ALG{UCA6} algorithm does (see
Line~\ref{lnl:uca6p-getsplittype} and Line~\ref{lnl:uca6p-choosethebest} in
Function~\ref{alg:split-uca6p}). The computed complementary boxes can be
memorized for improving the complementary boxing of subproblems. However, the
memorization should be made optional because it may make the computation slow.
Unlike the \ALG{UCA6} algorithm, the \ALG{UCA6$^+$} algorithm only memorizes
complementary boxes that do not contain the corresponding bounding box (see
Line~\ref{lnl:uca6p-remove-cb} in Algorithm~\ref{alg:uca6plus} and
Line~\ref{lnl:uca6p-remove-cb2} in Function \ref{alg:prunecheck-uca6p}).

Function \ref{alg:prunecheck-uca6p} (on page~\pageref{alg:prunecheck-uca6p})
attempts to apply a $\DRrd$ operator to the input subproblem in order to reduce
the domains of this subproblem. If it cannot prove that this subproblem is
infeasible, it then checks if the subproblem has at most $D_{\tn{stop}}$ active
variables. If the answer is yes, it resorts to a secondary solution technique,
called \ALG{DimStopSolver}, to solve the current subproblem, provided that
\ALG{DimStopSolver} provides an output with good alignments. A good candidate
for \ALG{DimStopSolver} is a search technique with the \emph{uniform cell
subdivision}\footnote{A \emph{uniform cell subdivision} means splitting the
domain box into equal $\varepsilon$-bounded boxes, called \emph{cells}. When
this splitting is used, the solver only need to solve subproblems defined on
each cell.} or the uniform bisection on all variables
\cite{SamHaroudD-F:1996,LottazC:Thesis:2000}. Variants of the \ALG{DMBC$^+$} or
\ALG{UCA6} algorithms that use the restricted-dimensional operators can also be
candidates.

Given an NCSP, $\ALG{DimStopSolver}$ constructs inner and boundary union
approximations of the solution set (see Line~\ref{lnl:uca6p-dimstopsolver} in
Function \ref{alg:prunecheck-uca6p}). These two union approximations are
naturally represented in DBR (or a bounding-box tree). They are converted to
EVR and then back to DBR in order to combine each group of contiguous aligned
boxes into a bigger equivalent box. This conversion procedure is performed by
the $\ALG{Combination}$ function.

\begin{theorem}
\label{thm:uca6p}%
Given a monotonic inclusion test $\FC$ and a positive precision (vector)
$\varepsilon$. The \ALG{UCA6$^+$} algorithm terminates and provides inner and
boundary union approximations, $\UNION{I}$ and $\UNION{B}$ respectively, at the
precision $\varepsilon$ with respect to the monotonic inclusion test $\FC$ (see
Definition~\ref{def:interval-precision}).
\end{theorem}

\begin{proof}
By an argument similar to the proof of Theorem~\ref{thm:uca5n6}, we have the
following properties:
\begin{enumerate}
    \item $\UNION{I}$ and $\UNION{B}$ are inner and boundary union approximations of
    the solution set, respectively; thus, $\UNION{I} \cup \UNION{B}$ is an
    outer union approximation of the solution set.

    \item If not applying the \ALG{Combination} function in
    Function~\ref{alg:prunecheck-uca6p}, every box $\mb{B}$ in $\UNION{B}$ can
    be split into $\varepsilon$-bounded boxes such that each resulting box
    $\mb{B}'$ satisfies the property: $\FC(\mb{B}', \cl{C}) = \CONST{unknown}$,
    where $\cl{C}$ is the set of running constraints in $\mb{B}$.

    \item Since the \ALG{Combination} function does not alter the union of
    boxes in $\UNION{B}$, the union of boxes in $\UNION{B}$ equals to the union
    of the above $\varepsilon$-bounded boxes.
\end{enumerate}
This implies what we have to prove.
\end{proof}

Notice that all the presented algorithms (\ALG{DMBC}, \ALG{DMBC$^+$},
\ALG{UCA5}, \ALG{UCA6} and \ALG{UCA6$^+$}) are complete if the inclusion test
$\FC$ is $\varepsilon$-strong for all sufficiently small $\varepsilon > 0$,
since they are all of the precision $\varepsilon$ w.r.t. $\FC$.

\section{Experiments}
\label{sec:search-experiment}%

Since all the above-presented search algorithms should work similarly and
equally for NCSPs with isolated solutions, in this paper we will only present
experiments on NCSPs with continuums of solutions. The set of benchmark
problems includes 14 nonlinear problems: $\mb{P1}$, $\mb{P2}$, $\mb{P3}$,
$\mb{P4}$, $\mb{FD}$, $\mb{G12}$, $\mb{H12}$, $\mb{F22}$, $\mb{L01}$,
$\mb{LE1}$, $\mb{S06}$, $\mb{S08}$, $\mb{TD}$, $\mb{WP}$. Their descriptions
are given in Appendix~\ref{sec:search-benchmarks}. They are NCSPs with
continuums of solutions that have been impartially chosen to show different
cases corresponding to different properties of constraints and solution sets
and that can be solved efficiently by at least one of the considered search
algorithms.

For the purpose of evaluation, we have implemented five search algorithms
(\ALG{DMBC}, \ALG{DMBC$^+$}, \ALG{UCA5}, \ALG{UCA6} and \ALG{UCA6$^+$}) with
different options using the same data structures and the same domain reduction
operators. Our experiments discarded \ALG{DMBC}, which is a point-wise
approach, as a reasonable candidate for solving NCSPs with continuums of
solutions because it usually produces a huge number of boxes, each is
$\varepsilon$-bounded, in very long running time. The source codes of the above
search algorithms can be found in the \PRO{BCS} 2.5.2 (\emph{box covering
solver}) module, which is downloadable at the official web site of the COCONUT
project, \url{http://www.mat.univie.ac.at/coconut-environment/}.

\begin{table}[!tb]
\vspace{-0.3em}%
\caption[The running time results for the search algorithms]{The running time
results for the search algorithms. The first seven problems are
three-dimensional while the last seven problems are two-dimensional.}
\label{tab:search-result-time}%
\renewcommand{\arraystretch}{1}
\setlength{\tabcolsep}{2.3pt}
\begin{footnotesize}
\begin{center}
\begin{tabular}[c]{|c||c|r|r|r|r|r|r|r|}
\hline
\begin{tabular}[c]{l}
    Algorithm \hfill $\blacktriangleright$
    \\
    Splitting type \hfill $\blacktriangleright$
    \\
    Memorization \hfill $\blacktriangleright$
\end{tabular}
& $\varepsilon$ &
\begin{tabular}[c]{c}
    \ALG{DMBC$^+$} \\ $\DS$ \\ No
\end{tabular}
&
\begin{tabular}[c]{c}
    \ALG{UCA6} \\ $\DS$ \\ Yes
\end{tabular}
&
\begin{tabular}[c]{c}
    \ALG{UCA6$^+$} \\ $\DS$ \\ No
\end{tabular}
&
\begin{tabular}[c]{c}
    \ALG{UCA5} \\ $\BS+\DS$ \\ No
\end{tabular}
&
\begin{tabular}[c]{c}
    \ALG{UCA6} \\ $\BS+\DS$ \\ Yes
\end{tabular}
&
\begin{tabular}[c]{c}
    \ALG{UCA6$^+$} \\ $\BS+\DS$ \\ No
\end{tabular}
&
\begin{tabular}[c]{c}
    Ratio \\ \shortstack[c]{$\frac{\mb{DMBC}^+}{\mb{UCA6}^+}$}
\end{tabular}
\\
\hline
\mb{P1}  & 0.1  & $>24$h    & 24.94s    & 19.34s    & 172.74s   & 3.76s    & 1.03s   & $>83883$  \\
\hline
\mb{P2}  & 0.1  & $>24$h    & 187.48s   & 95.70s    & 6.47s     & 7.80s    & 0.79s   & $>109367$ \\
\hline
\mb{P3}  & 0.1  & 37724.72s & 61.82s    & 25.40s    & 8.86s     & 14.23s   & 0.63s   & 59880     \\
\hline
\mb{P4}  & 0.1  & $>24$h    & 140.25s   & 92.89s    & 4.96s     & 4.51s    & 0.96s   & $>90000$  \\
\hline
\mb{FD}  & 0.1  & 505.77s   & 183.62s   & 101.29s   & 48.10s    & 59.13s   & 31.91s  & 15.8\\
\hline
\mb{G12} & 0.1  & 429.82s   & 172.52s   & 96.40s    & 32.72s    & 33.59s   & 22.23s  & 19.3     \\
\hline
\mb{H12} & 0.1  & 2161.17s   & 889.45s   & 267.36s   & 280.81s   & 273.64s  & 99.81s  & 21.7    \\
\hline
\mb{F22} & 0.01 & 5.14s    & 3.81s      & 3.98s     & 3.25s     & 3.50s    & 2.70s   & 1.9      \\
\hline
\mb{L01} & 0.01 & 2073.86s  & 1082.33s  & 660.74s  & 49.30s     & 51.08s    &  7.03s   & 295.0 \\
\hline
\mb{LE1} & 0.01 & 94.04s    & 39.79s    & 40.89s   & 34.35s     & 22.05s    &  7.32s   & 12.8 \\
\hline
\mb{S06} & 0.01 & 58.58s    & 44.29s    & 44.84s    & 29.78s    & 29.10s   & 24.34s   & 2.4     \\
\hline
\mb{S08} & 0.01 & 175.36s   & 89.25s    & 41.62s    & 10.05s    & 9.90s    & 5.72s    & 30.7    \\
\hline
\mb{TD}  & 0.01 & 9.82s     & 5.43s     & 6.64s     & 3.46s     & 3.82s    & 1.43s    & 6.9     \\
\hline
\mb{WP}  & 0.01  & 296.29s  & 85.82s    & 47.50s    & 26.20s    & 24.60s   & 17.21s   & 17.6    \\
\hline
\end{tabular}
\end{center}
\end{footnotesize}
\end{table}

\begin{table}[!htb]
\vspace{-1em}%
\caption[The numbers of boxes in inner and boundary union approximations]{The
numbers of boxes in inner union approximations (on the left) and boundary union
approximations (on the right).}
\label{tab:search-result-box}%
\renewcommand{\arraystretch}{1}
\setlength{\tabcolsep}{1.1pt}
\begin{footnotesize}
\begin{center}
\begin{tabular}[c]{|c|c|r|r|r|r|r|r|r|r|r|r|r|r|}
\hline
Prob. &
            & \multicolumn{2}{c}{\ALG{DMBC$^+$}} \vline%
            & \multicolumn{2}{c}{\ALG{UCA6}} \vline%
            & \multicolumn{2}{c}{\ALG{UCA6$^+$}} \vline%
            & \multicolumn{2}{c}{\ALG{UCA5}} \vline%
            & \multicolumn{2}{c}{\ALG{UCA6}} \vline%
            & \multicolumn{2}{c}{\ALG{UCA6$^+$}} \vline%
\\
$\blacktriangledown$ & $\varepsilon$
            & \multicolumn{2}{c}{$\DS$} \vline%
            & \multicolumn{2}{c}{$\DS$} \vline%
            & \multicolumn{2}{c}{$\DS$} \vline%
            & \multicolumn{2}{c}{$\BS+\DS$} \vline%
            & \multicolumn{2}{c}{$\BS+\DS$} \vline%
            & \multicolumn{2}{c}{$\BS+\DS$} \vline%
\\
$\blacktriangledown$ &
            & \multicolumn{2}{c}{Memo = No} \vline%
            & \multicolumn{2}{c}{Memo = Yes} \vline%
            & \multicolumn{2}{c}{Memo = No} \vline%
            & \multicolumn{2}{c}{Memo = No} \vline%
            & \multicolumn{2}{c}{Memo = Yes} \vline%
            & \multicolumn{2}{c}{Memo = No} \vline%
\\
\hline
\mb{P1}&0.1&$>$210000&$>$810000 & 10402 & 30601 & 10219 & 15716 & 63124 & 67824 & 4065  & 10854 & 785   & 1253  \\
\hline
\mb{P2}&0.1&$>$280000&$>$730000 & 21833 & 66223 & 15563 & 40027 & 8750  & 23920 & 8347  & 26643 & 523   & 1091  \\
\hline
\mb{P3}  & 0.1  & 106784& 528757& 8398  & 48080 & 5147  & 28038 & 10744 & 29812 & 11942 & 38502 & 369   & 932   \\
\hline
\mb{P4}&0.1 &$>$150000&$>$860000& 24230 & 62405 & 23901 & 31972 & 6643  & 13988 & 4979  & 13423 & 562   & 866   \\
\hline
\mb{FD}  & 0.1  & 16437 & 92681 & 16437 & 92681 & 15585 & 47990 & 51878 & 65536 & 26331 & 70218 & 10321 & 35134 \\
\hline
\mb{G12} & 0.1  & 17440 & 85062 & 17440 & 85062 & 12426 & 50878 & 34470 & 59440 & 24524 & 60526 & 13404 & 34590 \\
\hline
\mb{H12} & 0.1  & 27280 & 144296& 27280 & 144296& 18417 & 88212 & 75999 & 127436& 55080 & 127124& 29032 & 74656 \\
\hline
\mb{F22} & 0.01 & 1398  & 3458  & 1398  & 3458  & 1058  & 2260  & 1672  & 2584  & 1450  & 2664  & 906   & 1600  \\
\hline
\mb{L01} & 0.01 & 65705 & 106348& 65705 & 106348& 51838 & 67510 & 50031 & 67619 & 34296 & 67659 & 1857  & 2073  \\
\hline
\mb{LE1} & 0.01 & 14298 & 19688 & 14298 & 19688 & 9202  & 15331 & 13387 & 13795 & 8154  & 21918 & 1572  & 1496  \\
\hline
\mb{S06} & 0.01 & 12345 & 27756 & 12345 & 27756 & 8827  & 20154 & 23692 & 30439 & 11692 & 26008 & 9546  & 17486 \\
\hline
\mb{S08} & 0.01 & 26722 & 41208 & 26722 & 41208 & 16836 & 32478 & 15852 & 27384 & 15717 & 26624 & 9287  & 11716 \\
\hline
\mb{TD}  & 0.01 & 2881  & 3936  & 2881  & 3936  & 1936  & 2915  & 2685  & 4844  & 3160  & 4403  & 565   & 1091  \\
\hline
\mb{WP}  & 0.01 & 22212 & 38956 & 22212 & 38956 & 14341 & 29924 & 24465 & 36433 & 17264 & 33622 & 11273 & 18041 \\
\hline
\end{tabular}
\end{center}
\vspace{-0.2em}
\end{footnotesize}
\end{table}

\begin{table}[!htb]
\vspace{-0.5em}%
\caption[The ratios of the inner volume to the outer volume]{The ratios of the
volume of inner approximations to that of outer approximations.}
\label{tab:search-result-ratio}%
\vspace{0.8em}%
\renewcommand{\arraystretch}{1}
\setlength{\tabcolsep}{3.5pt}
\begin{footnotesize}
\begin{center}
\begin{tabular}[c]{|c||c|c|c|c|c|c|c|}
\hline
\begin{tabular}[c]{l}
    Algorithm \hfill $\blacktriangleright$
    \\
    Splitting type \hfill $\blacktriangleright$
    \\
    Memorization \hfill $\blacktriangleright$
\end{tabular}
& $\varepsilon$ &
\begin{tabular}[c]{c}
    \ALG{DMBC$^+$} \\ $\DS$ \\ No
\end{tabular}
&
\begin{tabular}[c]{c}
    \ALG{UCA6} \\ $\DS$ \\ Yes
\end{tabular}
&
\begin{tabular}[c]{c}
    \ALG{UCA6$^+$} \\ $\DS$ \\ No
\end{tabular}
&
\begin{tabular}[c]{c}
    \ALG{UCA5} \\ $\BS+\DS$ \\ No
\end{tabular}
&
\begin{tabular}[c]{c}
    \ALG{UCA6} \\ $\BS+\DS$ \\ Yes
\end{tabular}
&
\begin{tabular}[c]{c}
    \ALG{UCA6$^+$} \\ $\BS+\DS$ \\ No
\end{tabular}
\\
\hline
\mb{P1}   & 0.1  & n/a   & 0.980 & 0.979 & 0.997 & 0.997 & 0.990 \\
\hline
\mb{P2}   & 0.1  & n/a   & 0.972 & 0.967 & 0.996 & 0.996 & 0.985 \\
\hline
\mb{P3}   & 0.1  & 0.710 & 0.710 & 0.640 & 0.956 & 0.956 & 0.836 \\
\hline
\mb{P4}   & 0.1  & n/a   & 0.949 & 0.948 & 0.997 & 0.997 & 0.974 \\
\hline
\mb{FD}   & 0.1  & 0.984 & 0.984 & 0.983 & 0.992 & 0.992 & 0.986 \\
\hline
\mb{G12}  & 0.1  & 0.874 & 0.874 & 0.856 & 0.924 & 0.922 & 0.900 \\
\hline
\mb{H12}  & 0.1  & 0.885 & 0.885 & 0.868 & 0.938 & 0.937 & 0.918 \\
\hline
\mb{F22}  & 0.01 & 0.968 & 0.968 & 0.960 & 0.977 & 0.978 & 0.970 \\
\hline
\mb{L01}  & 0.01 & 0.999 & 0.999 & 0.999 & \AP 1 & \AP 1 & 0.999 \\
\hline
\mb{LE1}  & 0.01 & 0.997 & 0.997 & 0.995 & 0.999 & 0.999 & 0.997 \\
\hline
\mb{S06}  & 0.01 & \AP 1 & \AP 1 & \AP 1 & \AP 1 & \AP 1 & \AP 1 \\
\hline
\mb{S08}  & 0.01 & 0.999 & 0.999 & 0.999 & \AP 1 & \AP 1 & \AP 1 \\
\hline
\mb{TD}   & 0.01 & 0.996 & 0.996 & 0.995 & 0.998 & 0.999 & 0.995 \\
\hline
\mb{WP}   & 0.01 & 0.999 & 0.999 & 0.998 & 0.999 & 0.999 & 0.999 \\
\hline
\end{tabular}
\end{center}
\vspace{-0.5em}
\end{footnotesize}
\end{table}

In the above algorithms, the domain reduction operators ($\DR$ and $\DRrd$)
have been implemented using the function \texttt{IloGenerateBounds}, which is a
kind of domain reduction and a variant of box consistency, in a well-known
commercial product, \PRO{ILOG Solver} 6.0 \cite{ILOGSolver-6.0:2003}. The
complementary boxing operators ($\CB$ and $\CBrd$) have been implemented as in
Theorem~\ref{thm:dr-to-cb-operator}. The monotonic inclusion test $\FC$ has
been constructed as in Theorem~\ref{thm:fc-checker}. The inclusion test $\FC'$
in Function \ref{alg:prunecheckDMBCp} (Page~\pageref{alg:prunecheckDMBCp}) has
been implemented using a complementary operator, and thus returns either
$\CONST{feasible}$ or $\CONST{unknown}$. For simplicity, the experiments have
been taken with fixed settings for the new algorithms: \PRO{fragmentation
ratio} = 0.25, $D_{\tn{stop}} = 1$. All components of the vector $\varepsilon$
are assumed to be the same. The secondary search technique
(\ALG{DimStopSolver}) in the \ALG{UCA6$^+$} algorithm has been implemented as a
simple combination of a uniform cell subdivision and a monotonic inclusion
test.

The empirical results are presented in Table~\ref{tab:search-result-time},
Table~\ref{tab:search-result-box}, and Table~\ref{tab:search-result-ratio}.
Table~\ref{tab:search-result-time} shows the running time results of the
algorithms (in seconds and hours). Table~\ref{tab:search-result-box} shows the
numbers of boxes in inner and boundary union approximations delivered by the
algorithms. Table~\ref{tab:search-result-ratio} shows the ratios of the total
volume of inner union approximations to that of outer union approximations. The
term `Memorization' (Memo) indicates the memorization of complementary boxes
for the next iteration. The terms $\DS$ and $\BS+\DS$ in
Table~\ref{tab:search-result-time}, Table~\ref{tab:search-result-box}, and
Table~\ref{tab:search-result-ratio} indicate the splitting policies used in the
corresponding search algorithms:
\begin{itemize}
    \item $\DS$: always dichotomize the largest domain of the domain box.
    \item $\BS+\DS$: attempt to use a box splitting (\BS) first;
    if failed, then proceed with $\DS$.
\end{itemize}

Our experiments show that the new algorithms (\ALG{UCA5}, \ALG{UCA6} and
\ALG{UCA6$^+$}) is better than the classic algorithm (\ALG{DMBC$^+$}) in all
measures. The best gains of the new algorithms over the classic one are
obtained in case the arities of constraints are less than the arity of the
problem (e.g., four problems $\mb{P1}$--$\mb{P4}$). \emph{This shows how
important the reduction of arity of problems is}. In most cases, the \ALG{UCA5}
and \ALG{UCA6} algorithms with the option $\BS+\DS$ are quite equal in all
measures. However, the choice of constraints for splitting in \ALG{UCA6} is far
better than that in \ALG{UCA5} in the solution of $\mb{P1}$. The \ALG{UCA6$^+$}
algorithm with the option $\BS+\DS$ is always better than the others in the
running time and the number of boxes, even if it does not need the memorization
of complementary boxes (thus, less memory is need). The best gains of
\ALG{UCA6$^+$} over the others are obtained when constraint boundaries contain
a large percentage of nearly axis-parallel regions (e.g., $\mb{P2}$ and
$\mb{P3}$).

The \ALG{UCA6$^+$} algorithm with the option $\BS+\DS$ is slightly less
accurate than the \ALG{UCA5} and \ALG{UCA6} algorithms in the volume measure.
However, this situation get better when reducing $\varepsilon$. Moreover, this
is hardly a matter for real world applications, because no one could ever use
all solutions when a very large percentage of sound solutions has been found
and all the considered algorithms are of precision $\varepsilon$ w.r.t. $\FC$
(see Definition~\ref{def:interval-precision}). The \ALG{DMBC$^+$} algorithm and
the \ALG{UCA6} algorithm with the option $\DS$ produce similar outputs when all
constraints in a problem have the same set of variables, as happened for all
the problems except four problems, $\mb{P1}$--$\mb{P4}$. We observed that
\emph{the above gains of the new algorithms over the classic algorithm,
\ALG{DMBC$^+$}, get better when reducing $\varepsilon$, especially for hard
problems}.

Notice that the arities of constraints in all the above problems, except
$\mb{P1}$--$\mb{P4}$, equal to the arities of the problems. In fact, only the
experiments on $\mb{P1}$--$\mb{P4}$ may show the full effectiveness of the new
algorithms, including the reduction of the arity of problem during the solution
(we recall that the time and space complexities of the algorithms are
exponential in the arity of problem). The experiments on the other problems do
not show the the same improvements as those on $\mb{P1}$--$\mb{P4}$. This
reveals that the effect of the arity reduction during the solution in new
algorithm is an important improvement. Other experiments also show a similar
relation among the search algorithms using a variant of hull consistency in
\cite{BenhamouF-G-G-P:1999}.

\section{Conclusion}
\label{sec:search-conclusion}%

In this paper, we presented a uniform view on search strategies of
branch-and-prune methods for solving NCSPs. In this view, we started with a
generic branch-and-prune algorithm, \ALG{BnPSearch}, and then derived from it
two classic algorithms, \ALG{DMBC} and \ALG{DMBC$^+$}, for the point-wise and
set-covering approaches, respectively. As the main contribution of the paper,
we proposed three new branch-and-prune search algorithms: \ALG{UCA5},
\ALG{UCA6} and \ALG{UCA6$^+$}. Presenting the new algorithms as instances of
\ALG{BnPSearch} and extensions of \ALG{DMBC$^+$} facilitates the comparison of
them. In particular, we clearly presented the differences among the algorithms.

Our experiments show that the \ALG{UCA6$^+$} algorithm (with the option
$\BS+\DS$) is the most adaptive search, in time and compactness, among the
search algorithms for NCSPs with continuums of solutions, while the \ALG{UCA5}
and \ALG{UCA6} algorithms (with the option $\BS+\DS$) seem to be able to
balance between speed and accuracy in most cases. They are all far better than
the classic branch-and-bound search algorithms, \ALG{DMBC} and \ALG{DMBC$^+$},
especially when the arities of constraints are less than the arity of the
problem. Moveover, the new algorithms often provide a large percentage of sound
solutions (in the form of a collection or tree of inner boxes) when solving
NCSPs with continuums of solutions.

In case the solution set consists of continuums but is highly disconnected, one
may wish to cluster its union approximations to get grouping/clustering
information on them. In this case, we propose to use the clustering techniques
\cite{VuXH-SH-F:2003a} to perform a post-processing on the approximations to
generate such information.

We predict that the \ALG{UCA6$^+$} algorithm will show more speed up and
compactness if we use higher values, $D_{\tn{stop}} = 2, 3$ and use the search
technique in \cite{SamHaroudD-F:1996,LottazC:Thesis:2000} in place of
\ALG{DimStopSolver}. A direction for further research is to explore different
combinations of pruning techniques and other tests such as existence,
uniqueness, exclusion, and inclusion tests to make new branch-and-prune search
methods, especially for addressing different classes of problems. Comparisons
with a broader range of search algorithms are also needed.

\begin{acks}

Support for this research was partially provided by the European Commission and
the Swiss Federal Education and Science Office (OFES) through the COCONUT
project (IST-2000-26063). We would like to thank ILOG for the software licenses
of ILOG Solver used in the COCONUT project, and Professor Arnold Neumaier at
the University of Vienna (Austria) for fruitful discussions and very valuable
input.

\end{acks}


\renewcommand{\refname}{\footnotesize REFERENCES}
\bibliographystyle{acmtrans}

\newpage
\appendix

\section{Numerical Benchmarks}
\label{sec:search-benchmarks}%

\subsection{Problem \mb{TD}}

\begin{figure}[!htb]
\centering
    \includegraphics[width=9cm]{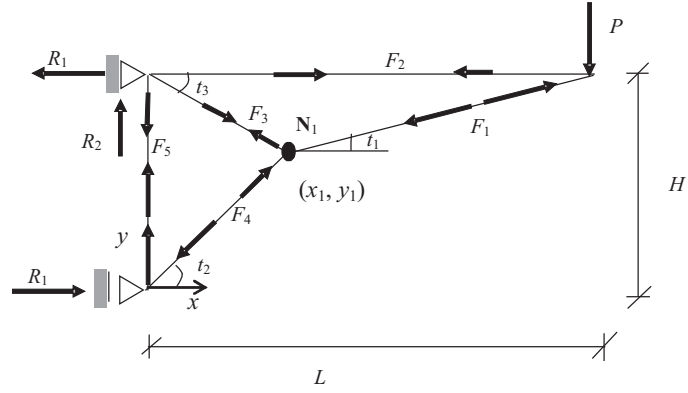}
\caption[The geometric design of a truss]{The geometric design of a truss.}
\label{fig:truss-design}%
\end{figure}

Consider the geometric design problem of a truss depicted in
Figure~\ref{fig:truss-design}. The goal is to find the coordinates of the
moveable joint $(x_1, y_1)$ in $\cintv{0.01}{10} \times \cintv{0.01}{10}$ of
the node $\mb{N}_1$ of the truss such that all the following constraints are
satisfied:
{\allowdisplaybreaks%
\begin{eqnarray*}
    F_2 & < & T A,
    \\
    F_5 & < & T A,
    \\
    F_1 & < & C_1 A,
    \\
    F_1 & < & T A,
    \\
    F_4 & < & C_4 A,
    \\
    F_4 & < & T A,
    \\
    |F_3| & < & T A,
    \\
    F_3 \le 0 & \OIF & -F_3 < C_3 A,
    \\
    x_1 & < & L,
    \\
    y_1 & < & H,
\end{eqnarray*}
}%
where
{\allowdisplaybreaks%
\begin{equation*}
    \begin{array}{rcll}
        E & = & 210 * 10^6 & \tn{Young's modulus of steel, unit = kN/m$^2$};
        \\
        T & = & 235 * 10^3 & \tn{The yield stress of steel, unit = kN/m$^2$};
        \\
        A & = & 0.25 & \tn{The area of cross section of truss members};
        \\
        r & = & 0.5 & \tn{The radius of gyration of the cross section of truss
        members};
        \\
        P & = & 400 & \tn{The loading capacity};
        \\
        H & = & 6 & \tn{The height of truss};
        \\
        L & = & 10 & \tn{The length of truss};
    \end{array}
\end{equation*}}
and the auxiliary variables are defined as follows:
{\allowdisplaybreaks%
\begin{eqnarray*}
    \tan t_1 & = & (H - y_1) / (L - x_1),
    \\
    \tan t_2 & = & y_1 / x_1,
    \\
    \tan t_3 & = & (H - y_1) / x_1,
    \\
    R_1 & = & P L / H,
    \\
    F_1 & = & P / \sin t_1,
    \\
    F_2 & = & P / \tan t_1,
    \\
    F_3 & = & (R_1 - F_2) / \cos t_3,
    \\
    F_4 & = & R_1 / \cos t_2,
    \\
    F_5 & = & R_1 \tan t_2,
    \\
    L_1 & = & \sqrt{(L - x_1)^2 + (H - y_1)^2},
    \\
    L_3 & = & \sqrt{x_1^2 + (H - y_1)^2},
    \\
    L_4 & = & \sqrt{x_1^2 + y_1^2},
    \\
    C_1 & = & \pi^2 E /(L_1 / r)^2,
    \\
    C_3 & = & \pi^2 E/(L_3 / r)^2,
    \\
    C_4 & = & \pi^2 E/(L_4 / r)^2.
\end{eqnarray*}
}%
In fact, this is a two-dimensional problem: the variables are $x_1$ and $y_1$.
All the other variables can be easily eliminated in a preprocessing phase. The
reduced constraints are however too complicated (in the number of elementary
operations) to be read, and is hence not listed here.

\subsection{Problem \mb{FD}}

Consider the design problem of the beam of a railway bridge under cyclic
stress. The goal is to find $(L, q_f, Z) \in [10, 30] \times [70, 90] \times
[0.1, 10]$ such that the following yield stress and fatigue stress are
satisfied:\footnote{The variable $Z$ is scaled up 100 times in unit in
comparison to the original version.}
\begin{eqnarray*}
    \sigma & < & f_y,
    \\
    \sigma_e & < & \tn{resistance},
\end{eqnarray*}
where
{\allowdisplaybreaks%
\begin{equation*}
    \begin{array}{rcll}
        \sigma_c & = & 115000 & \tn{Yield stress of steel, unit = kN/m$^2$};
        \\
        \gamma & = & 1.1 & \tn{The safety factor};
        \\
        f_y & = & 460000 & \tn{Unit = kN/m$^2$};
        \\
        \tn{years} & = & 200 & \tn{The number of years to fatigue failure};
    \end{array}
\end{equation*}
}%
and the auxiliary variables are defined as follows:
\begin{equation*}
    \alpha = \left\lbrace
    \begin{array}{ll}
        1.3 & \tn{ if } L \le 4,
        \\
        1.3 - 0.1 (L - 4) & \tn{ if } 4 < L \le 7.5,
        \\
        0.95 - 0.008 (L - 7.5) & \tn{ if } 7.5 < L \le 20,
        \\
        0.85 - (L - 20)/300 & \tn{ if } 20 < L \le 50,
        \\
        0.75 & \tn{ if } L > 50,
    \end{array}
    \right.
\end{equation*}
{\allowdisplaybreaks%
\begin{eqnarray*}
    \phi & = &  0.82 + 1.44 / (\sqrt{L} - 0.2),
    \\
    q_r & = & q_f \phi,
    \\
    \sigma & = & q_r  L^2 / 8 / (Z/100),
    \\
    \sigma_e & = & \alpha \sigma,
    \\
    \tn{cycles} & = &  0.05\ \tn{years},
    \\
    \sigma_r & = & \sigma_c (\min\set{2.5, \tn{cycles} / 2})^{-1/3},
    \\
    \tn{resistance} & = & \sigma_r/\gamma,
\end{eqnarray*}
}%

\subsection{Problem \mb{WP}}

This is a two-dimensional simplification of the design model for a kinematic
pair consisting of a wheel and a pawl. The constraints determine the regions
where the pawl can touch the wheel without blocking its motion.
\begin{equation*}
    \left\lbrace
    \begin{array}{ll}
        20 < \sqrt{x^2 + y^2} < 50;
        \\
        12y/\sqrt{(x-12)^{2}+y^{2}} < 10;
        \\
        x \in \sintv{50},\ y \in \cintv{0}{50}.
    \end{array}
    \right.
\end{equation*}

\subsection{Problem \mb{P1}}

Three dimensions; the arities of constraints are less than the arity of
problem:
\begin{equation*}
    \left\lbrace
    \begin{array}{ll}
        2x^{2} \le 3y - (y + 1)^{0.2} + 5;
        \\
        \ln (y^{3/2} + 2y + 1) + 5 \le z + (z + 1/2)^{0.1};
        \\
        (x + 1)^{1.5} \ge 2 \sqrt{x} / (3 + \sqrt{z^2 + 1});
        \\
        x \in \cintv{0}{50},\ y \in \cintv{0}{100},\ z \in \cintv{0}{50}.
    \end{array}
    \right.
\end{equation*}

\subsection{Problem \mb{P2}}

Three dimensions; the arities of constraints are less than the arity of
problem:
\begin{equation*}
    \left\lbrace
    \begin{array}{ll}
        x^{2} \le y;
        \\
        \ln y + 1 \ge z;
        \\
        xz \le 1;
        \\
        x \in \cintv{0}{15},\ y \in \cintv{1}{200},\ z \in \cintv{-10}{10}.
    \end{array}
    \right.
\end{equation*}

\subsection{Problem \mb{P3}}

\textbf{P2} added with the fourth constraint whose arity equals to the
problem's arity:
\begin{equation*}
    \left\lbrace
    \begin{array}{ll}
        x^{2} \le y;
        \\
        \ln y + 1 \ge z;
        \\
        xz \le 1;
        \\
        x^{3/2} + \ln(1.5z + 1) \le y+1;
        \\
        x \in \cintv{0}{15},\ y \in \cintv{1}{200},\ z \in \cintv{0}{10}.
    \end{array}
    \right.
\end{equation*}

\subsection{Problem \mb{P4}}

Three dimensions; the arities of constraints are less than the arity of
problem:
\begin{equation*}
    \left\lbrace
    \begin{array}{ll}
        x^{1.5} + 1.9 \le \ln(y^3 + y + 1.5);
        \\
        \ln(y^2 + z + 1) \le z + 2;
        \\
        \sqrt{x^2 + z^2 + 12x + 5} \le 3 + (2x + 3)^3;
        \\
        x \in \cintv{0}{50},\ y \in \cintv{0}{100},\ z \in \cintv{0}{50}.
    \end{array}
    \right.
\end{equation*}

\subsection{Problem \mb{G12}}

Three dimensions; the arities of constraints are equal to the arity of problem:
\begin{equation*}
    \left\lbrace
    \begin{array}{lll}
    x_1^2 + 0.5 x_2 + 2 (x_3 - 3) \ge 0;
    \\
    x_1^2 + x_2^2 + x_3^2 \le 25;
    \\
    x_1, x_2, x_3 \in \sintv{8}.
    \end{array}
    \right.
\end{equation*}

\subsection{Problem \mb{H12}}

Three dimensions; the arities of constraints are equal to the arity of problem:
\begin{equation*}
    \left\lbrace
    \begin{array}{lll}
    x_1^2 + x_2^2 + x_3^2 \le 36;
    \\
    (x_1 - 1)^2 + (x_2 - 2)^2 + x_3^2 \ge 16;
    \\
    x_1^2 + (x_2 - 0.4)^2 \ge 2 x_3;
    \\
    x_1, x_2, x_3 \in \sintv{10}.
    \end{array}
    \right.
\end{equation*}

\subsection{Problem \mb{F22}}

Two dimensions; the intersection of a tricuspoid and a circle:
\begin{equation*}
    \left\lbrace
    \begin{array}{lll}
    (x^2 + y^2 + 24 x + 36)^2 \le 64 (x + 3)^3;
    \\
    x^2 + y^2 \ge 8;
    \\
    x, y \in \sintv{4}.
    \end{array}
    \right.
\end{equation*}

\subsection{Problem \mb{L01}}

Two dimensions; a problem with logarithm and power operations:
\begin{equation*}
    \left\lbrace
    \begin{array}{lll}
    (x + 0.1) \sqrt{y} \ge 20 + \sqrt{x};
    \\
    \ln (\sqrt{y + 1} + 13) +  50 \ge (x + 0.5)^{1.2};
    \\
    x \in \cintv{0}{50},\ y \in \cintv{0}{200}.
    \end{array}
    \right.
\end{equation*}

\subsection{Problem \mb{LE1}}

Two dimensions; a problem with logarithm, square root and exponent operations:
\begin{equation*}
    \left\lbrace
    \begin{array}{lll}
    e^{x + 1} / e^{\sqrt{y + 1}} \le  100 \sqrt{x y + 7} + 30;
    \\
    (x^2 - 3 x + 1) \sqrt{y + 2} \ge x \ln(10 y + 3) + 50;
    \\
    x, y  \in \cintv{0}{50}.
    \end{array}
    \right.
\end{equation*}

\subsection{Problem \mb{S06}}

Two dimensions; a single constraint whose solution set consists of disconnected
subsets:
\begin{equation*}
    \left\lbrace
    \begin{array}{lll}
    12 y / \sqrt{(x-12)^2 + y^2} \le 10;
    \\
    x \in \sintv{50},\ y  \in \cintv{0}{50}.
    \end{array}
    \right.
\end{equation*}

\subsection{Problem \mb{S08}}

Two dimensions; the difference between two circles with interior:
\begin{equation*}
    \left\lbrace
    \begin{array}{lll}
    20 \le \sqrt{x^2 + y^2} \le 50;
    \\
    x \in \sintv{50},\ y  \in \cintv{0}{50}.
    \end{array}
    \right.
\end{equation*}

\endreceived%
\end{document}